\documentclass[letterpaper, 10 pt, conference]{ieeeconf}  % Comment this line out if you need a4paper
\renewcommand{\baselinestretch}{0.999} 

\IEEEoverridecommandlockouts                              % This command is only needed if 
\usepackage[utf8]{inputenc} % allow utf-8 input
\usepackage[T1]{fontenc}    % use 8-bit T1 fonts
\usepackage{xr}
\usepackage{hyperref}       % hyperlinks
\usepackage{url}            % simple URL typesetting
\usepackage{booktabs}       % professional-quality tables
\usepackage{amsfonts}       % blackboard math symbols
\usepackage{nicefrac}       % compact symbols for 1/2, etc.
\usepackage{microtype}      % microtypography

\usepackage{graphicx}
\usepackage{subcaption}

\usepackage{amsthm}
\usepackage{amsmath, amssymb}
\usepackage{bbm}
\usepackage{bm}
\usepackage{siunitx}
\usepackage{dsfont}
\newtheorem{lemma}{Lemma}

\newtheorem{theorem}{Theorem}
\usepackage[ruled,linesnumbered]{algorithm2e}

\title{\LARGE \bf
Adaptive Variance for Changing Sparse-Reward Environments
}

\author{Xingyu Lin$^{1}$, Pengsheng Guo$^{1}$, Carlos Florensa$^{2}$, David Held$^{1}$\\
  $^{1}$ Robotics Institute, Carnegie Mellon University\\
  $^{2}$ EECS Department, UC Berkeley\\}

\begin{document}

\maketitle
\thispagestyle{empty}
\pagestyle{empty}

%%%%%%%%%%%%%%%%%%%%%%%%%%%%%%%%%%%%%%%%%%%%%%%%%%%%%%%%%%%%%%%%%%%%%%%%%%%%%%%%

\begin{abstract}
Robots that are trained to perform a task in a fixed environment often fail when facing unexpected changes to the environment due to a lack of exploration. We propose a principled way to adapt the policy for better exploration in changing sparse-reward environments. Unlike previous works which explicitly model  environmental changes, we analyze the relationship between the value function and the optimal exploration for a Gaussian-parameterized policy and show that our theory leads to an effective strategy for adjusting the variance of the policy, enabling fast adapt to changes in a variety of sparse-reward environments.
\end{abstract}

\section{Introduction}
\begin{figure}[t]
\centering
\includegraphics[width=0.45\linewidth]{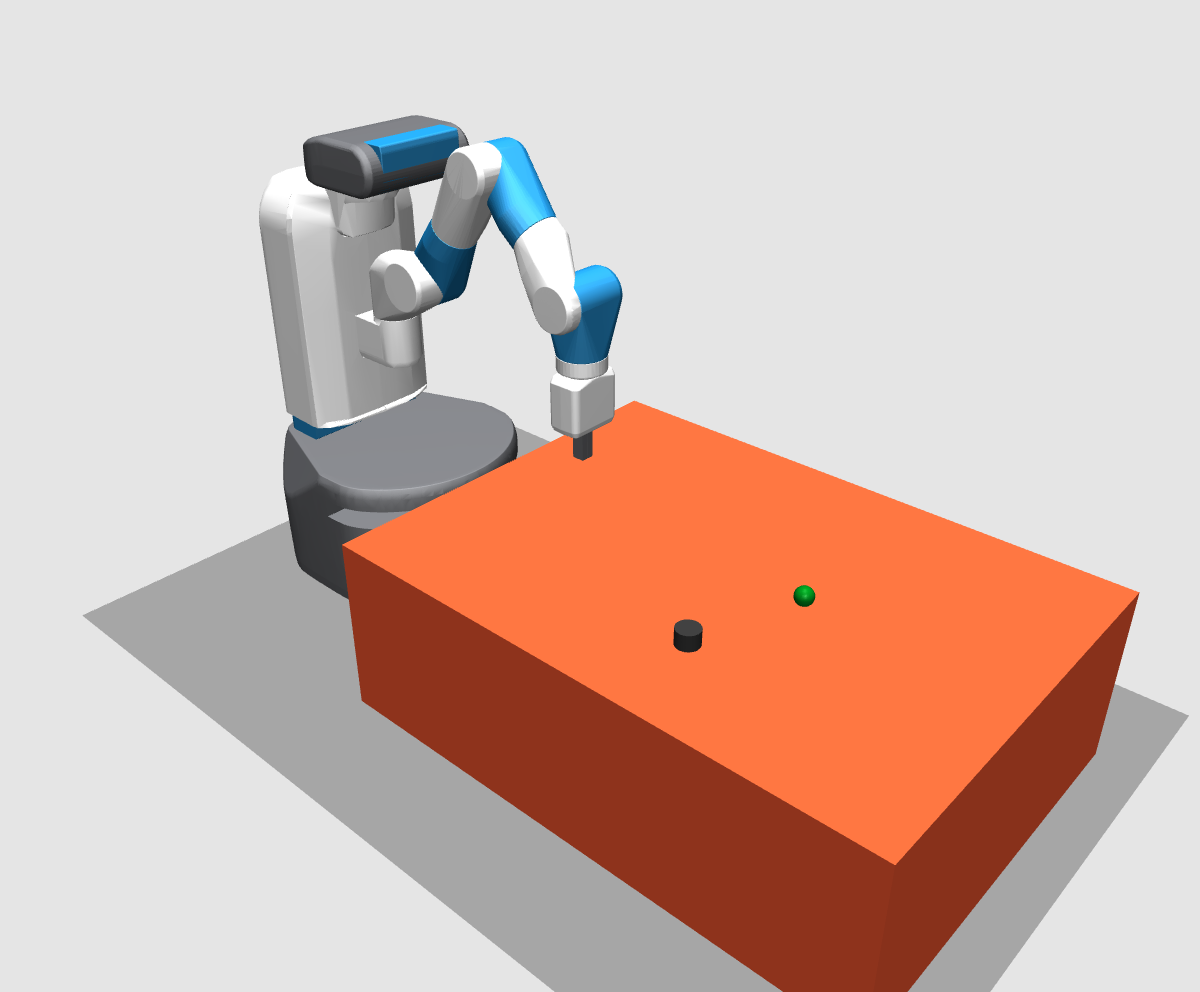}
\includegraphics[width=0.45\linewidth]{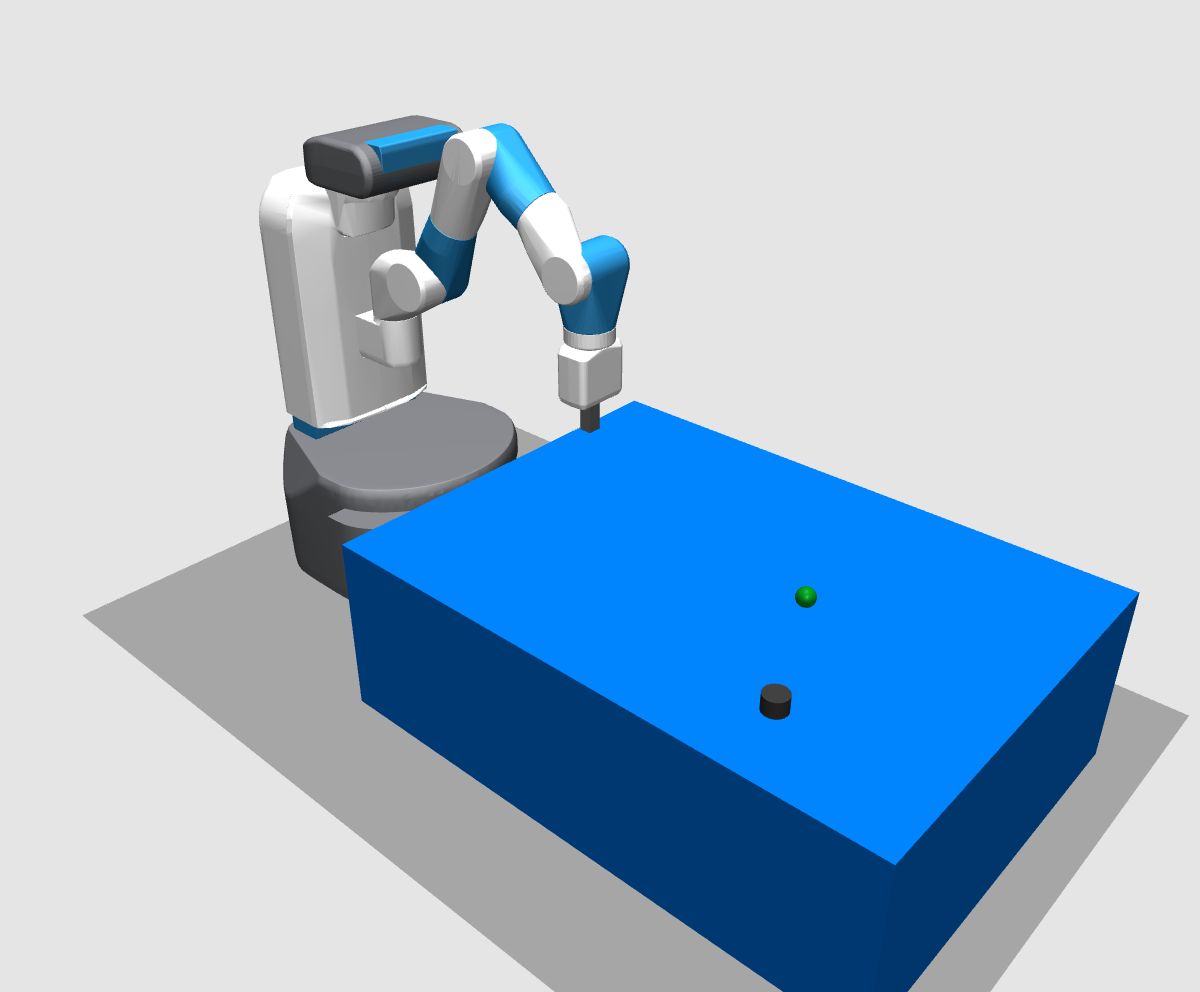}
\caption{Left: The robot is trained to push a block to a target position on a given table surface.  Right: Later, the robot is placed in front of a different surface which might (or might not) have a different coefficient of friction.  Our method enables the robot to quickly adapt its policy to such changes in the environment.}
\label{fig:pull}
\end{figure}

Reinforcement learning has demonstrated great potential in a variety of different robotics tasks, such as teaching a humanoid to stand up or run~\cite{schulman2017proximal}, or learning dexterous manipulation skills \cite{Andrychowicz2018, rajeswaran2017learning}.  However, all of these environments share the property that the reward function and the dynamics model defining these environments are fixed.  On the other hand, a robot must be able to adapt to unexpected changes in its environment.   For example, if a robot is trained to push objects on a table, the friction coefficient between the objects and the table may change over time (see Figure~\ref{fig:pull}), due to wearing down of either the object or the table through repeated use. A robot that is navigating outdoors might transition from navigating on grass to concrete. Objects in the environment that were previously located in one position may get moved to a different position. Finally, if a robot is trained in simulation, it will later need to transfer its knowledge to the real world, which may have different dynamics, due to unmodeled effects or incorrectly estimated parameters. This problem is commonly known as ``distributional shift" and has been identified as one of they key research directions for AI progress and safety \cite{Amodei2016AIsafety}. The specific question that we want to investigate is: how should the robot explore to optimally adapt to environmental changes?

%a manipulation robot might get bumped by an external force resulting in a discrete change in the robot’s position.  

%A robot that is trained to interact with one set of objects may need to interact later with a different set of objects.  

%The environment that an agent interacts with can change due to a number of different causes. 

%the friction or damping of a robot can change over time.
% I removed this because the reviewers last time asked why we are mostly considering abrupt changes as opposed to smooth ones

In reinforcement learning, the agent must sample actions in order to decide how much to update its policy.  The actions are sampled from an exploration policy, which can be the same as the policy itself (for on-policy methods) or it might be different (for off-policy methods).  In either case, the exploration policy includes parameters that relate to how much exploration the agent will undergo.  One open question is how an agent should choose these parameters that guide how much exploration it should perform.

For agents that learn via exploration, the reinforcement learning paradigm presents a problem for environments that are sparse or that have many local optima: without good exploration, the agent has a low probability of sampling good actions which would help it determine how to optimally update its exploration parameters.  The agent is caught in a vicious cycle in which its poor exploration leads to a poor update of its exploration parameters, and the agent is not able to properly adapt its policy.

%For agents that learn via exploration, the reinforcement learning paradigm presents a problem: without good exploration, the agent cannot discover good actions and thus it cannot determine how to optimally update its exploration parameters.  The agent is caught in a vicious cycle in which its poor exploration leads to a poor update of its exploration parameters, and the agent is not able to adapt.

% \begin{figure}[t]
% \centering
% \includegraphics[width=0.9\linewidth]{figures/intro.png}
% \caption{An illustration of the distribution shift in the environment. The agent might decrease the amount of exploration for high performance in the first stage and fail to explore in the second stage.}
% \label{fig:intro}
% \end{figure}

We investigate this challenge for a particular class of problems in which an agent in a continuous action space receives sparse rewards. For this class of problems, we propose a method that allows an agent to quickly adapt to changes in its environment dynamics.  After the agent has learned to successfully master a task in one environment, the environment dynamics or the reward function might change. Our insight is that, after training to convergence in a fixed environment, the value function of the agent in a binary sparse reward setting (in which the undiscounted return ranges from 0 to 1, and in which a return of 1 is feasible from every initial state) gives us a measure of how much the environment has shifted:  a value of 1 corresponds to no shift and a value of 0 corresponds to a drastic shift.  Based on the value, the agent can estimate the amount of environmental shift and thereby compute how much exploration is necessary, leading to  significantly faster adaptation than previous approaches. 

%We analytically derive the relationship between value and variance for optimal exploration, and we show that this method enables our agent to achieve significantly faster adaptation than previous approaches. 

%  

%We show that the amount of exploration that should be performed can be computed analytically as a function of the value of the agent. 

Our contributions in this paper are as follows:
\begin{itemize}
%\item We propose a principle approach to adjust the variance in Gaussian-parameterized policies in response to environmental changes in tasks with sparse rewards

%to balance the exploration and exploitation, which is robust to sudden environment changes without modeling the changes.

%\item We analytically compute the relationship between the amount of distribution shift, the sparsity of the environment, and the optimal exploration for a Gaussian-parameterized policy

\item We theoretically analyze the relationship between the value function and the optimal exploration for a Gaussian-parameterized policy in a sparse reward environment

\item We use the above computation to propose a principled approach to adjust the variance of a  Gaussian-parameterized policy in response to environmental changes in tasks with sparse rewards

\item We demonstrate empirically that our proposed approach provides a practical solution that allows a policy to adapt quickly to environmental changes in a variety of different sparse-reward environments, including robot manipulation tasks.

%performs well in a variety of sparse-reward environments and environmental changes.

%\item We give a solution to the value-variance mapping functions analytically, as well as a practical solution that performs well in a variety of environments.
\end{itemize}

\section{Related Work}
The problem of distributional shift has been extensively studied in the supervised learning literature \cite{quinonero2009shift}, often under the restrictive assumption of ``covariate shift'' \cite{sugiyama2012covariate}. Often, it is studied from an Active Learning perspective \cite{littlestone1988learning}, and under some conditions the number of errors can be  bounded \cite{Yang2011drifting,Li2008kwik}. However, these results do not extend to the Reinforcement Learning setup.

%(i.e. access to actions taken by the optimal policy at that state)
%, potentially making them more attractive
%might help to recover from a change in the environment, but they 

% The below paragraph is interesting but not really necessary to include.

%In most Reinforcement Learning algorithms there is an intrinsic ``covariate shift" because the behavior policy used to collect data changes through training iterations, hence affecting the observed state-distribution. This is necessary because of the lack of supervised labels of the optimal action in each state, and hence exploration is needed to find the best action given the state. Nevertheless, most of these exploration strategies fail to recover when an extrinsic distributional shift happens, such as a modification of the reward function or the environment dynamics. 

%Indeed, 

In reinforcement learning, distribution shift presents a problem because the agent must sufficiently explore to discover the optimal policy.  Most current methods fail to recover when an extrinsic distributional shift happens, such as a modification of the reward function or the environment dynamics, since many exploration strategies converge towards a purely exploiting policy. For example, bootstrapped DQN \cite{Osband2016bootstrapped} which learns multiple Q functions for better exploration, tends to have all its Q functions coincide when a static task is mastered -- hence losing the uncertainty estimate if the environment changes; the same is true for soft-Q learning and other entropy regularized methods \cite{Haarnoja2017energy, Fox2016noise, Nachum2017pcl}. For count-based 
bonuses strategies \cite{tang2017exploration, Bellemare2016count} the exploration bonus converges to zero for states that were previously visited sufficiently often, so there would be no incentive to re-visit those states even if the environment changed. Another approach is to use a learned dynamics model to provide an exploration bonus~\cite{Houthooft2016vime, Achiam2017surprise}; although such an approach can theoretically help with distribution shift, our experiments demonstrate that this approach is fairly unstable, whereas our much simpler approach reliably leads our agent to recover from environmental changes.  There has been some work on finding optimal exploration strategies for discrete action spaces~\cite{ODonoghue2017uncertainty, Tokic2010adaptive,JMLR:v17:14-335,sakaguchi2004reliability}, but such methods do not easily transfer to continuous action spaces that we are investigating.

Another approach that has recently been investigated is to train the agent in a sufficiently diverse setting such that no exploration is needed at test time. These methods assume that the agent can observe (or sample from) the set of all possible environments in advance (e.g. in simulation).  Then the robot can just identify which of the previously observed environments it is encountering, either explicitly~\cite{Yu2017unknown} or implicitly, using a latent representation~\cite{Peng2018randomization, Rajeswaran2016epopt,finn2017maml,Duan2016rl2,Al-Shedivat2017continuous}. However, the assumption that our learning algorithm has access to the distribution of all possible environments in training time is not realistic.  First, the number of parameters that describe our environment may be very large, and training our method to be robust to all possible combinations of such parameters will take an exponentially long time.  Second, the environment may change in unpredictable ways that we did not anticipate in training time.  In contrast, our method adapts online and can handle unexpected environmental changes that were not anticipated in advance. 

The problem of exploration under uncertain environments can also be formulated as solving a non-stationary MDP. Many works in this area assume that the non-stationary MDP consist a number of unknown stationary MDP, such as Hidden-Mode MDP\cite{choi2000hidden}, which assumes a fix number of modes, or construct partial models for new MDP on the fly \cite{da2006dealing}. These methods try to explicitly predict the mode changes and learn a new model accordingly. In contrast, our method predicts the environment changes implicitly and in a continuous way and is able to utilize the previously learned skills.

In this work, we propose a value-dependent exploration strategy for fast adaptation to environmental changes in sparse reward environments. This idea was previously mentioned in \cite{gullapalli1990stochastic} without any theoretical justification. In contrast, we analytically compute the relationship between the value and the optimal variance show that our results leads to improved performance in different robotic tasks. 

%This identification can be done explicitly, by predicting a known set of environmental parameters \cite{Yu2017unknown} or implicitly, using a latent representation \cite{Peng2018randomization, Rajeswaran2016epopt}. Another line of work with similar assumptions is meta-learning~\cite{finn2017maml, Duan2016rl2}, where the agent is trained to quickly adapt by gathering new data ~\cite{Al-Shedivat2017continuous}.
%}, even in non-stationary environments \cite{

%seen all possible environments in training time is not realistic.  

\section{Methodology}
\subsection{Problem Definition}
We start with a discrete-time finite-horizon Markov descision process (MDP) denoted by a tuple $(\mathcal{S}, \mathcal{A}, \mathcal{P}, r, \gamma, \rho_0, T)$, where $\mathcal{S}$ is a state set, $\mathcal{A}$ is an action set, $\mathcal{P}:\mathcal{S} \times \mathcal{A} \times \mathcal{S} \rightarrow \mathbb{R}_+$ is the transition probability distribution, $r: \mathcal{S}\rightarrow \mathbb{R}_+$ is the reward function, $\rho_0:\mathcal{S} \rightarrow \mathbb{R}_+$ is the distribution of the initial state $s_0$, $\gamma \in [0,1]$ is the discount factor and $T$ is the time horizon. A stochastic policy is defined as $\pi: \mathcal{S} \times \mathcal{A} \rightarrow [0, 1]$. The value function is defined by the accumulated, discounted future return at state $s$, follow policy $\pi$: $V_\pi(s) = \mathbb{E}_\pi[\Sigma_{i=t+1}^{\infty}\gamma^{i-t-1} R_i | S_t=s]$. Similarly, the Q function is defined as: $Q_\pi(s, a) = \mathbb{E}_\pi[\Sigma_{i=t+1}^{\infty}\gamma^{i-t-1} R_i | S_t=s, A_t=a]$.

Most RL algorithms with an explicit policy over a continuous action space parameterize the policy as a diagonal Gaussian distribution~\cite{schulman2015trpo,mnih2016asynchronous}. The mean is typically a parameterized function of the state, $a \sim \mathcal{N}(\bm{\mu}, \Sigma)$ The variance $\Sigma = diag\{\sigma_i\}$, is usually either also parameterized as a function of the state \cite{degris2012model} or defined as a global parameter (i.e.\ independent of the state)~\cite{schulman2015trpo}.  A common approach is to learn the variance along with the mean with a policy gradient method; however, this approach leads to the policy converging to deterministic for a fixed environment, which is a poor outcome if the environment changes at some point in the future. Alternatively, one can choose a fixed variance; however, choosing a good fixed variance is hard: a variance too large will hurt performance as the policy cannot reliably execute precise actions, while a variance too small is not sufficient for exploration. 

We instead propose to learn the variance as a function of the value of the current state. Intuitively, if the agent is in a state where it has a low chance of getting a high reward, we should use a large variance to increase the level of exploration. On the other hand, we should use a small variance when in a state from which our policy has already learned to achieve a high reward, to best exploit this learned knowledge. Next, we will formalize this intuition.
% two conditions: 1) sparse reward at the end 2) dense reward

%define the ``success'' of the agent in the
%as a measure of the agent's ``success" (defined as the normalized reward)

%\textbf{Single-reward environment} 

\subsection{Sparse Reward Environments} \label{sec:Success Prediction}
We consider a single-reward environment, where we only get a single reward $r(s_T) \in [R_{min}, R_{max}]$ at the last time step of an episode, usually denoting whether the task is completed successfully. The episode is terminated after the reward is given, so the undiscounted return for a trajectory $\tau$ also ranges from $r(\tau) \in [R_{min}, R_{max}]$. $R_{min}$ and $R_{max}$ are fixed for an environment, and we assume that there exists a trajectory $\tau$ starting from each state $s$ that has support in the initial state distribution $\rho_0$ such that $r(\tau \mid s_0 = s) = R_{max}$. In this way, we can always normalize the undiscounted return and map it onto the range $[0, 1]$, where 0 is the lowest possible return and 1 is the largest possible return (which is assumed to be achievable from every initial state).  The return can thus be interpreted as the fraction of the maximum return that the current agent can achieve, giving a measure of how much it can still improve.

%We then define the agent's ``success" for a given episode as the value of this normalized reward received at the end of the episode. The success can also be interpreted as the fraction of the maximum reward that the current agent achieves, giving a measure of how much it can still improve. This can be seen as a generalization of the goal-oriented environments defined in \cite{florensa2017reverse, held2017goal}. In the special case of the reward being binary (either $R_{min}$ or $R_{max}$), the success can also be interpreted as the probability of reaching the maximum reward.

%, summarizing the performance of the trajectory $\tau_i$ in this episode

We train an undiscounted value function $V_\phi:S\mapsto [0,1]$ that, given the current state, predicts the expected undiscounted normalized return of the agent starting from that state. The value function is trained to minimize the loss function $$\frac{1}{NT}\sum_{i=1}^N\sum_{t=1}^T||V_\phi(s^{(i)}_t) - r^{(i)}_T ||_2^2,$$
where $r_T$ is the undiscounted normalized return, which is equal to the reward received at the end of the episode. In practice, the learned value function $V_\phi(s)$ is clipped onto the range $[0, 1]$.

\subsection{Value Dependent Exploration} \label{sd_exploration}
We then learn a function $f_\rho\colon[0,1]\mapsto \mathbb{R}_+^{|\mathcal{A}|}$ that maps from the undiscounted value function $V_\phi(s)$ to the desired action variance at that state $s$. This mapping allows the agent to modulate its exploration solely based on its level of mastery at that state. Potentially the agent could learn to do more exploration in un-mastered states (where the value is low) while performing less exploration in mastered states (where the value is high). Unlike exploration strategies that are based on visitation counts \cite{tang2017exploration,Bellemare2016count}, the value dependent exploration will keep exploring at states where it may have visited many times but still have a poor performance, while exploiting states that it has already mastered even if the state was only visited a few times. 

% We expect that the mapping $f$ will generalize well to new environmental parameters, allowing the agent to quickly adapt to environmental changes (and our experiments demonstrate this to be true). 
Below, we will derive the optimal form that the mapping function $f_\rho$ should take.  We will first derive this function for the continuous bandit case, and then we will generalize our result to the MDP case.  Due to the strong prior induced by the form of these expressions, the mapping function can be learned fairly easily and generalize well to new environmental parameters, allowing the agent to quickly adapt to new situations. 

%, allowing the agent to adapt faster to new situations and generalize better across environments.
%the monotonicity of the variance mapping function in the context
% and we get the reward based on the action we took
% either fixed or learned by gradient descent

\textbf{Optimal Variance in Continuous Bandit.} Intuitively, we may want to use a large action variance for exploration when in an un-mastered state and do less exploration when in a mastered state. We now formalize this intuition for the case of a continuous bandit problem \cite{agrawal1995continuum}.  The environment in such problems is constrained such that there is only one state and the episode ends after one time step. Because the action space is continuous, we assume that the policy is parameterized by a Gaussian distribution $a \sim \mathcal{N}(\mu,\,\sigma^{2})$, where $\mu = \pi(s)$ and $\Sigma = diag\{\sigma_i\}$. In the following sections we will explore how to choose the variance $\Sigma$ for robustness to environment changes and better exploration. We define the sparse reward function at time $t$ as 
\[ r_t(a)=
    \begin{cases} 
        1 & l(t) \leq a \leq l(t) + w \\
        0 & \textrm{otherwise},
    \end{cases}
\]
where $w$ is the (unknown) constant that determine the width of the interval in the action space from which we get a positive reward, and $[l(t), l(t)+w]$ defines the region in action space where a positive reward is given.  Define $d=l(t)-\mu$ as the distance from the action mean to the closest action that can get a positive reward which also changes with $t$. Since all the variables dependent on $t$ can be viewed as indirectly dependent on $t$ through $d$, we will write all the variables dependent on $d$ instead of $t$. For example, we use $V_\pi(\sigma,d)$ instead of $V^t_\pi(\sigma)$. $\mu$ is the action mean which is kept fixed during our analysis. We can then write the value function of a policy as a function of $\sigma$ and $d$: 
%; $d$ is also related to the amount of distribution shift
%For the remainder of this analysis, we consider optimization of the variance $\sigma$ in one iteration and fix $\mu$ and $d$. Thus, we can write the value function of a policy as a function of $\sigma$: 
\begin{equation}
V_\pi(\sigma, d) = \mathbb{E}_{a \sim \mathcal{N}(\mu,\,\sigma^{2})}[r_t(a)] \label{equ_v_expectation}
\end{equation}
With a fixed $\mu$, the optimal variance to use for the policy is the one that maximizes the value function: 
$$ \sigma^*(d) = \operatorname*{argmax}_\sigma V_\pi(\sigma, d)$$
The proof of all of the below lemmas and theorems can be found in Appendix A and B \footnote{The appendix and other supplementary materials can be found at \href{https://sites.google.com/andrew.cmu.edu/adaptive-variance}{https://sites.google.com/andrew.cmu.edu/adaptive-variance}}.

First, we determine the optimal value for the variance, given full knowledge of the environment:
\begin{lemma} \label{sigma_best}
Under the condition that $ w, d, \sigma>0 $, the optimal variance is given by
\begin{equation}
\sigma^*=\sqrt{\frac{1}{2} \frac{2dw+w^2}{\ln(1+w/d)}}.
\label{eq:lemma1}
\end{equation}
\end{lemma}
In Appendix D, we empirically validate the correctness of Lemma \ref{sigma_best} by showing that empirically optimizing the reward gives the same variance as computed in Equation~\ref{eq:lemma1}. However, in order to compute $\sigma^*$ using Lemma~\ref{sigma_best}, we have to first know the distance $d$. Instead of explicitly estimating $d$, we next relate $\sigma^*$ to  $V_\pi(\sigma^*)$. % When $V_\pi(\sigma^*)$ decreases, assuming $w$ is fixed, we can infer how $d$ changes and thus infer how $\sigma^*$ changes.
\begin{theorem} \label{theorem_inverse}
Under the condition that $ w, d, \sigma>0 $ and $d \gg w$, the optimal variance can be written in terms of the value function as
\begin{equation} \label{eq:v-inverse}
\sigma^* = \frac{w}{\sqrt{2\pi e} V_\pi(\sigma^*, d)}
\end{equation}
\end{theorem}
In practice, we can approximate $V_\pi(\sigma^*, d)$ with a Monte Carlo estimate of Equation \ref{equ_v_expectation}.  Note that this approximation will be biased because the variance that we use for sampling might be different from the optimal variance.  The parameter
 $w$ in equation~\ref{eq:v-inverse} is unknown and is jointly trained with the rest of the policy parameters.  On the other hand, the dependency on $d$ was removed since this value can change greatly due to a distribution shift, so we can adapt faster by removing the dependence on this variable.
 
 %Since we are interested in the relationship between the action variance and the value function at the current state $s_t$ which is fixed, 

\textbf{Monotonic Variance Mapping in MDP.} 
Generalization of our results from the continuous bandit case to the MDP case is challenging since the shape of the Q function is harder to model than the reward function. While we can no longer derive a closed form solution to the optimal variance, we can still show that the optimal function mapping from the value to the variance is monotonically decreasing.
For notational convenience, we omit $s_t$ in the following analysis (for example, the $Q$ function will be written as a function of only action $a$). First, we assume that the initial Q function (before any environment changes) is a bounded function in the action dimension, i.e.
$$ Q_0(a) = 0 \text{ if } a < l_0   \text{ or } a > r_0;\\$$
in other words, $Q_0(a)$ is only non-zero for $l_0 < a < r_0$.  Additionally,  we assume that all the rewards are non-negative and thus $\forall a, Q_0(a)\geq0$. Define $w= r_0 - l_0$. As the environment changes, the Q function will also change. Here we assume that when the change in the environment is relatively small, the Q function changes only through a shift in the action space. Given $d$ as the distance from $\mu$ to the closest point where Q is non-zero, we can define $l(d) = \mu + d$ and $r(d) = \mu+d+w$ for the Q function $Q_d$ after an environmental change. As $Q_d$ is (by assumption) only a translation of $Q_0$, we have that $Q_d(a) = Q_0(a-(l(d)-l_0))$. 
By definition of the value function, we have $$V_\pi(d) = \int \pi(a) Q_d(a)da,$$ where $V_\pi(d)$ is the value function given a certain $d$. Since the policy is parameterized as a normal distribution $\mathcal{N}(\mu, \sigma)$ and we assume $\mu$ is fixed, we can write the value function as a function of $\sigma$ and $d$: $V(\sigma, d) = V_\pi(d)= \int p(a| \mu, \sigma) Q_d(a),$ where $p(\cdot | \mu, \sigma)$ is the density of the Gaussian distribution.

\begin{lemma} \label{mdp_prop_sigma_bound}
Under the condition that $d, w, \sigma >0$, if the mean action of the policy $\mu$ is fixed, the optimal variance $\sigma^* \in (d, d + w)$. 
\end{lemma}
From Lemma \ref{mdp_prop_sigma_bound}, we can easily see that, if $d$ increases at least by $w$, then $\sigma^*$ should also increase by at least $w$. This holds true without any constraints on $w$ and $d$. Furthermore, a stronger monotonicity relationship is given by the following theorem:
\begin{theorem} \label{mdp_theorem_monotonic}
Under the condition that $d, w, \sigma >0$, if $w$ and $d$ satisfies that $w/d < \sqrt{3} -1$ then the optimal variance $\sigma^*$ increases when $V(\sigma^*, d)$ decreases.
\end{theorem}
This theorem allows us to learn a monotonic mapping function from the estimated value function to the optimal variance. Our assumption that the Q-function is bounded can be easily relaxed to the case where the $(1-\alpha)$-highest density region~\cite{hyndman1996computing} of the Q function is bounded. In this case, we  assume that most of the actions that achieve a non-zero Q-function can be grouped within a small region and we can get a guarantee of near-optimality for an approximate optimal variance $\hat{\sigma}^*$, where $|V(\sigma^*, d) - V(\hat{\sigma}^*, d)| < 1-\alpha$. The detailed proof can be found in Appendix B.

% The above theorems in both the continuous bandit case and the MDP case allow us to choose the policy variance to maximize the value function.  We further relate the value function to the speed of learning for batch policy updates: %when episodes are grouped together for a batch policy update:
% \begin{prop} \label{prop_positive_reward}
% When using policy gradient methods with a batch size of $n$, assuming $V_\pi(\sigma)$ is close to zero, if we increase the value function $V_\pi(\sigma)$ by $\delta$, the probability of getting at least one positive reward within the batch increases by $n \delta$.
% \end{prop}

%For environments where the reward is either 0 or 1, w

%%%%%%%%%%%%%%%%%%%%%%%%%%%%%%%%%%%%%%%%%%%%%%%%%%%%%%%%%%
% \begin{algorithm}[t]
% \caption{Success dependent exploration}\label{framework}
% $\pi_{\theta_0}$: Policy.\;
% $V_{\theta_1}$: Success predictor.\;
% $f_{\theta_2}$: Function that maps from predicted success to variance.\;
% $Iter$: Training iterations.
% \begin{algorithmic}[1]
% \For{$i \gets 1 \textrm{ to } Iter$}
%         Collect trajectories with \;
%         \qquad $a_t \sim \mathcal{N}(\pi_{\theta_2}(s_t), f_{\theta_2}(V_{\theta_1}(s_t)))$
%         \quad Update $\pi_{\theta_0}, f_{\theta_2}$ with policy gradient methods \label{line:1}
%         \quad Update $V_{\theta_1}$ with sampled trajectories \label{line:2}
% \EndFor
% \end{algorithmic}
% \end{algorithm}
%%%%%%%%%%%%%%%%%%%%%%%%%%%%%%END%%%%%%%%%%%%%%%%%%%%%%%%%%%%

\textbf{Practical algorithm for MDP}
A practical framework of our algorithm is shown in Algorithm \ref{framework}. We choose TRPO~\cite{schulman2015trpo} for our policy gradient updates in Line \ref{line:1}. We need to make sure to update the policy before updating the value function, as shown in Lines \ref{line:1} and Line \ref{line:2}. While this means that our policy gradient is calculated based on the old value function, it ensures that the KL divergence constraint used in TRPO is not violated purely due to the changes in the value function and variance.

\begin{algorithm}
\DontPrintSemicolon
$\pi_{\theta}$: Policy.\;
$V_{\phi}$: Undiscounted value function.\;
$f_{\rho}$: Function that maps from value to variance.\;
$Iter$: Training iterations. \;
\For{$i \gets 1 \textrm{ to } Iter$}
{
Rollout with $a_t \sim \mathcal{N}(\pi_\theta(s_t), f_\rho(V_\phi(s_t)))$ \;
Update $\pi_\theta, f_\rho$ with policy gradient methods \label{line:1}\;
Update $V_\phi$ with sampled trajectories \label{line:2}\;
}
\caption{Value dependent exploration}\label{framework}
\end{algorithm}

% We choose TRPO~\cite{schulman2015trpo} for our policy gradient updates in Line \ref{line:1}. Note that in Line \ref{line:1} and \ref{line:2}, we update the policy before updating the predicted success. While this means that our policy gradient is calculated based on the old predicted success, it makes sure that the KL divergence constraint used in TRPO is not violated purely due to the changes in predicted success and variance.
\begin{figure}[h]
\centering
\includegraphics[width=0.95\linewidth]{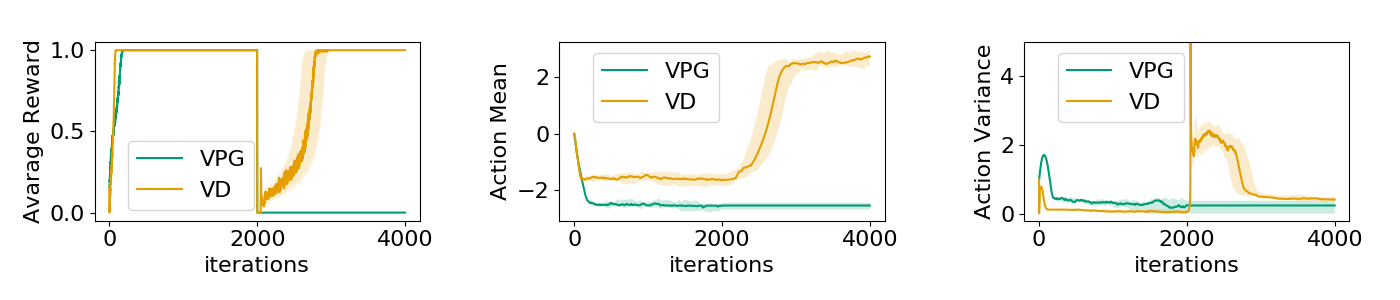}
\caption{Average reward (left), action mean (middle), and variance (right) of the vanilla policy gradient (VPG) and our value dependent exploration (VD) on the continuous bandit example.}
\label{fig:simple}
\end{figure}

\begin{figure*}
    \centering
    \begin{subfigure}[t]{0.9\linewidth}
        \includegraphics[width=\textwidth]{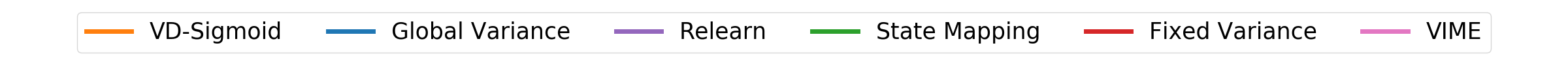}
    \end{subfigure}
    \begin{subfigure}[t]{0.24\linewidth}
        \includegraphics[width=\textwidth]{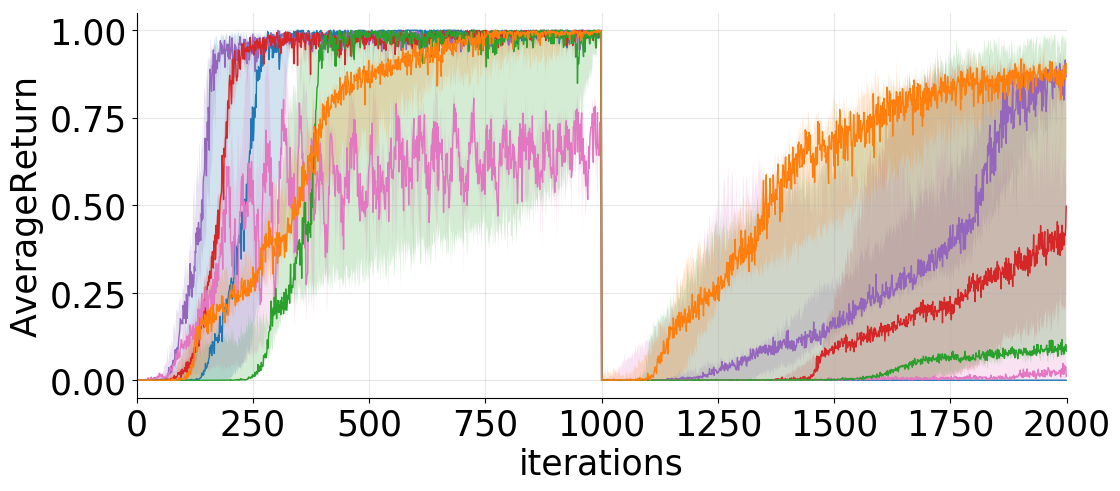}
        \caption{DIP-center.} 
        \label{fig:dip-center-learning}
    \end{subfigure}
    \begin{subfigure}[t]{0.24\linewidth}
        \includegraphics[width=\linewidth]{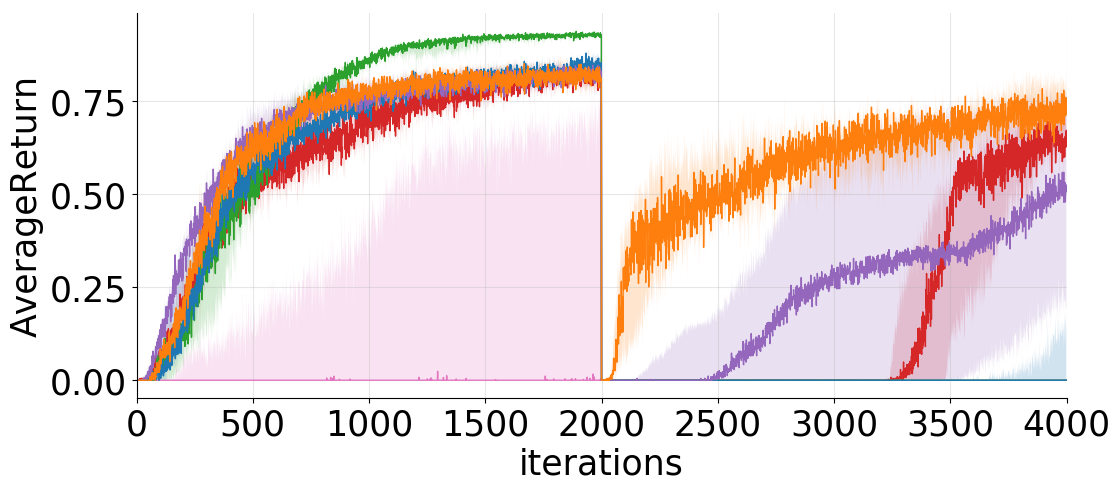}
        \caption{Thrower.} 
        \label{fig:thrower-learning}
    \end{subfigure}
    \begin{subfigure}[t]{0.24\linewidth}
        \includegraphics[width=\linewidth]{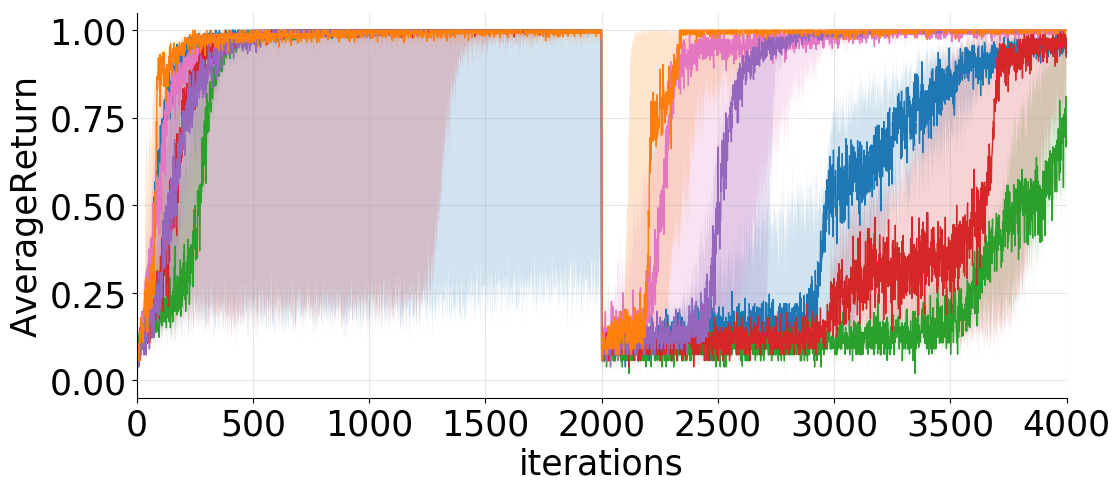}
        \caption{BiC.} 
        \label{fig:ball_in_cup_learning}
    \end{subfigure}
    \begin{subfigure}[t]{0.24\linewidth}
        \includegraphics[width=\linewidth]{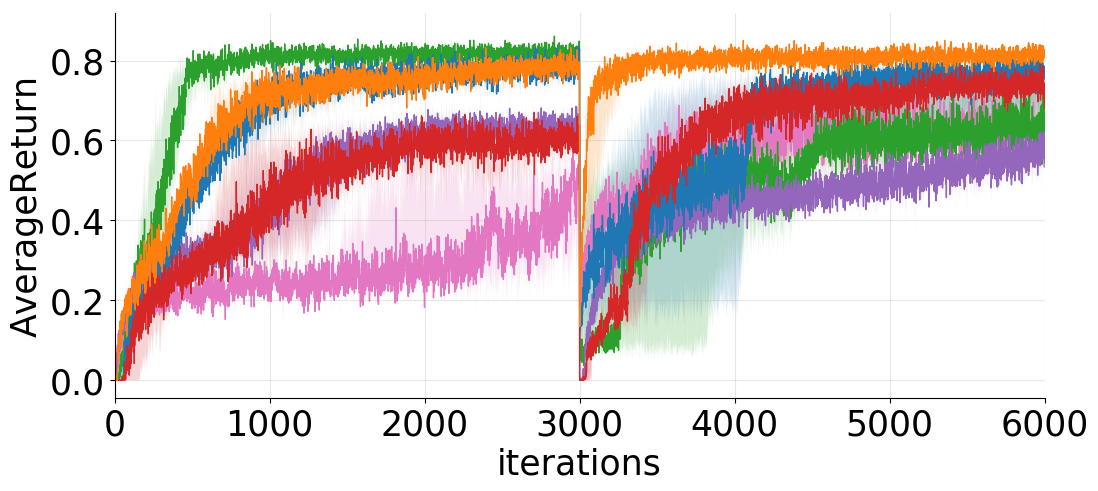}
        \caption{Slide.} 
        \label{fig:fetch_slide_learning}
    \end{subfigure}
    \caption{Learning curves on a variety environments and different environment changes.}\label{fig:all_experiments}
\end{figure*}

\begin{figure*}
    \centering
    \begin{subfigure}[t]{0.24\linewidth}
        \includegraphics[width=\textwidth]{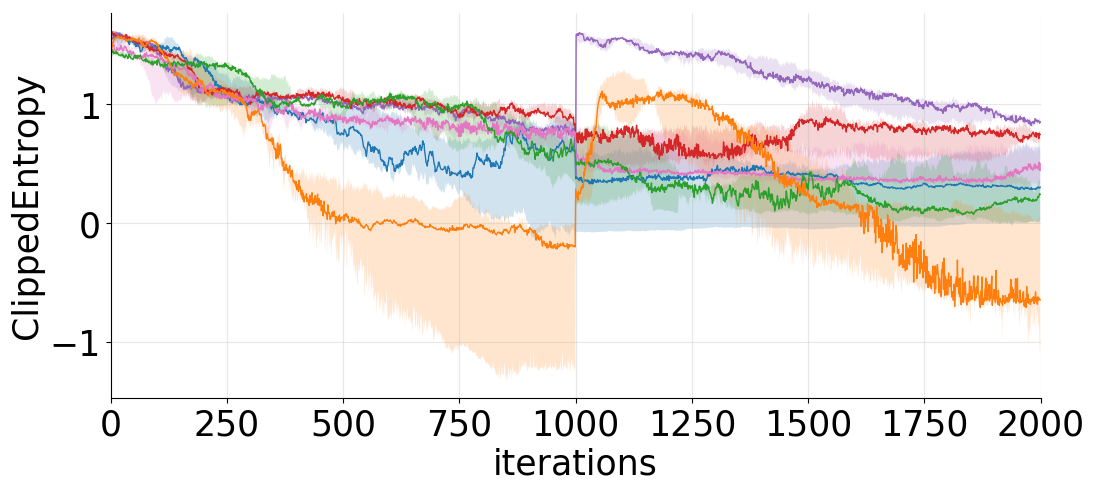}
        \caption{DIP-center.}
        \label{fig:dip-center-entropy}
    \end{subfigure}
    \begin{subfigure}[t]{0.24\linewidth}
        \includegraphics[width=\linewidth]{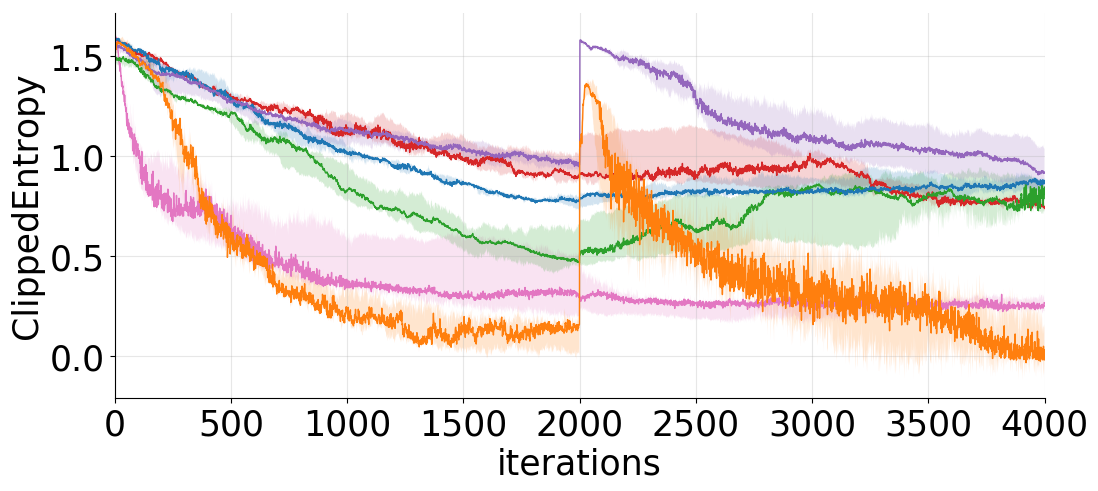}
        \caption{Thrower.}
        \label{fig:thrower-entropy}
    \end{subfigure}
    \begin{subfigure}[t]{0.24\linewidth}
        \includegraphics[width=\linewidth]{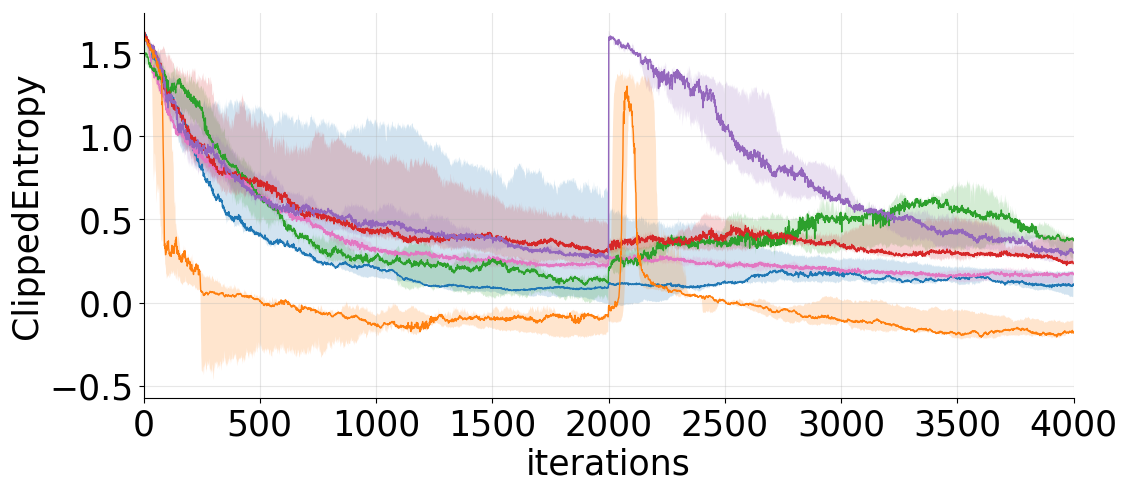}
        \caption{BiC.}
        \label{fig:ball_in_cup_entropy}
    \end{subfigure}
    \begin{subfigure}[t]{0.24\linewidth}
        \includegraphics[width=\linewidth]{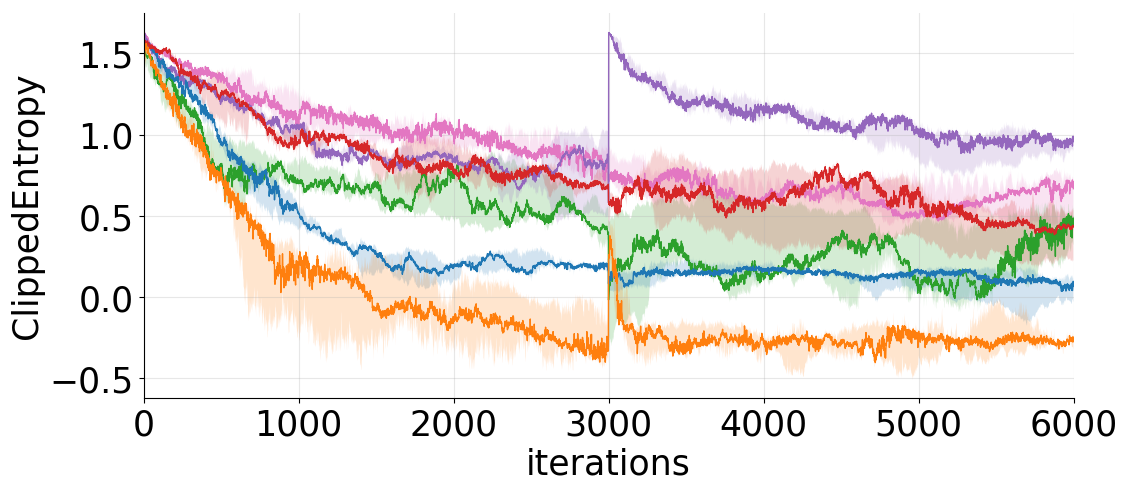}
        \caption{Slide.}
        \label{fig:fetch_slide_entropy}
    \end{subfigure}
    \caption{Average clipped entropy on a variety environments and different environment changes.}\label{fig:all_experiments_entropy}
\end{figure*}

\section{Evaluation}
We evaluate our method on situations where there is a distribution shift in the environment. Specifically, the environments consist of two stages and there is a change in the environment parameters between the two stages, such as the position of objects, their orientation, or their center of mass (i.e. new objects that the agent encounters with a different mass distribution). In all our experiments, we plot the color band representing the 25 percentile and 75 percentile. %, although the some curves have very low variance and the distribution is difficult to visualize.

\subsection{Continuous bandit example}
We first test our method in a continuous bandit setting, with a one dimensional action space, which satisfies the assumptions stated in Section \ref{sd_exploration}. The reward is defined as 
\[ r_t(a)=
    \begin{cases} 
        \mathds{1}\{a \in [-10, -1]\}, & t \in [1, 2000] \quad~~\text{(Stage I)} \\
        \mathds{1}\{a \in [1, 10]\}, & t \in[2001, 4000]  \text{(Stage II)}
    \end{cases}
\]
The policy is initialized as $a \sim \mathcal{N}(0, 1)$. We compare two methods: 1) Vanilla policy gradient (REINFORCE \cite{williams1992reinforce}), where both the policy mean and variance are learned by gradient descent, and 2) Value dependent exploration, where the variance $\sigma_i = \frac{w}{\sqrt{2\pi e} \hat{V}_i}$, as suggested by Theorem \ref{theorem_inverse}. The parameter $w$ is learned by gradient descent. In both cases, the policy gradient update does not use a baseline function.  We use ADAM \cite{kingma2014adam} for adjusting the learning rate for both methods. The result is shown in Figure \ref{fig:simple}. We can see that before the environment changes at $t=2000$, both methods learn quickly, while the value dependent exploration decreases the variance faster to exploit better. When the environment changes, the vanilla policy gradient is not able to learn anymore due to the lack of positive reward and the limited exploration. On the other hand, the value dependent exploration quickly increases its variance when the value function drops, enabling the policy to receive positive rewards and learn much faster.

We provide further analysis on the convergence of the learned variance to the optimal variance in Appendix D, as well as the convergence of $w$ to the true value, showing that we can learn the optimal mapping from value to variance by gradient descent, despite the approximations.

\subsection{Manipulation Environments}
We further evaluate our method on complex manipulation tasks simulated in Mujoco~\cite{todorov2012mujoco}. We first specify a training iteration for the first stage so that all algorithms converge. Then we enter the second stage and change the environment in some way, without signaling the algorithms. We use TRPO for updating the policy mean. We use a policy modeled by a (32, 32) multi-layer perceptron (MLP) for all the algorithms. We have a value function $V_{discounted}$ with a discounted factor of 0.99 for baseline estimation, which is a (32, 32) MLP. The undiscounted value function $V_\phi$ that provides the input for our variance mapping function is of the same size. In all the experiments each dimension of the action space is normalized to $[-1, 1]$ and the actions output by the policy are also clipped to be within this range. 

%In practice, in order to provide more degrees of freedom to the mapping function, we use the following form
%mapping function in the form of Equation \ref{sd-sigmoid}, denoted as 

For our value dependent exploration, we use a mapping function of the form:
\begin{equation} \label{sd-sigmoid}
\hat{\sigma}^*(s) = \frac{max(a,0)}{e^{k(V_{\phi}(s)-b) }+1} + max(c,0),
\end{equation}
where $k,a,b,c$ are parameters that we can train with policy gradient updates. 
We denote this method as VD-Sigmoid (``VD" stands for ``value dependent"). This function is guaranteed to be monotonically decreasing, as suggested by Theorem \ref{mdp_theorem_monotonic}. However, this form is more flexible than the inverse function used for the continuous bandit case in Equation~\ref{eq:v-inverse}. Empirically, we found this parameterization of the mapping function performs well across the tasks. In all our experiments, the initial values of the parameters $k, a, b$, and $c$ are set to be $5, 1.2, 0.3$, and $0.1$ respectively. Examples of the learned variance mapping functions can be found in Appendix D.

%(and these parameters are jointly trained along with our policy parameters via policy gradient updates).

%The algorithm using Equation \ref{eq:v-inverse}, denoted as SD-inverse, is compared to in Section ~\ref{ablation}.

\textbf{Baselines} We compare our method with four baselines which adjust the variance in different ways: 
\begin{itemize}
    \item Fixed variance: The variance is fixed to be $1$.  This value was empirically determined to perform well, when the action is normalized to be in the range of $[-1, 1]$. 
    \item Global variance: The variance is defined as a global parameter, which is updated by policy gradient updates at each iteration, as was done in~\cite{schulman2015trpo}.
    \item State mapping: A neural network is learned to map each state to a variance, where the network is updated by policy gradient updates, as was done in \cite{degris2012model}.
    \item VIME~\cite{Houthooft2016vime}: VIME explores by maximizing the information gain computed by the posterior of a learned dynamics model.
    \item Relearn: Assume that drift detection \cite{Gama2004detection} or context detection \cite{da2006dealing} is done perfectly, i.e. we know when the environment changes, a model is trained from scratch when the environment changes \cite{Gama2004detection}. The baseline receives additional information of when the environment changes. We use a fixed variance for this baseline. 
\end{itemize}

\begin{figure*}[h!]
    \centering
    \begin{subfigure}[h]{0.24\linewidth}
        \includegraphics[width=\linewidth]{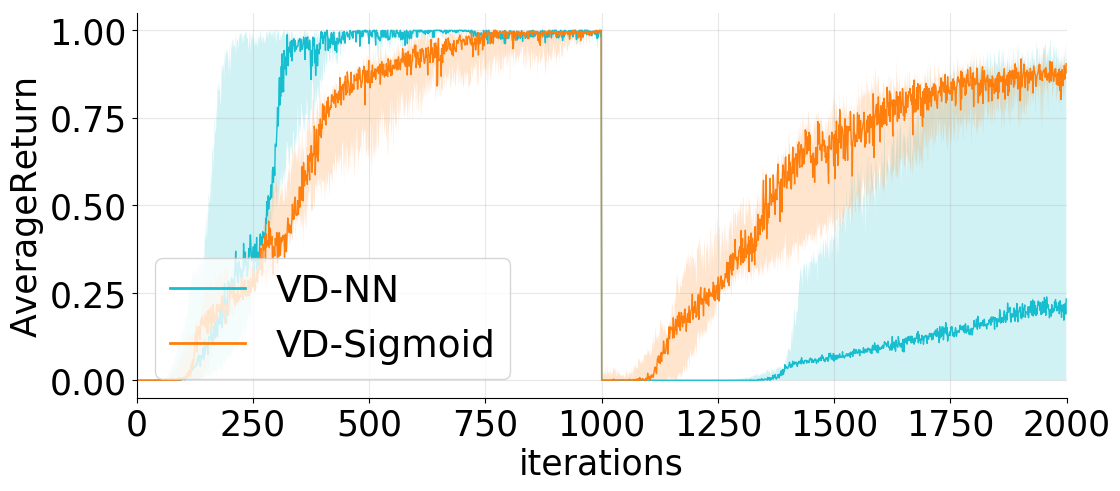}
        \caption{DIP-center.}
        \label{fig:ab-dip-center-auc}
    \end{subfigure}
    \begin{subfigure}[h]{0.24\linewidth}
            \includegraphics[width=\linewidth]{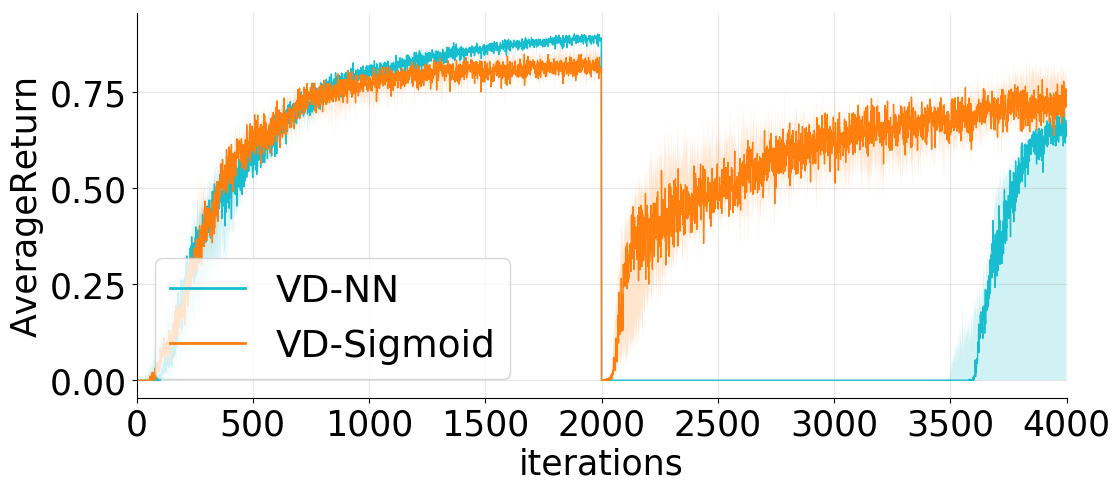}
        \caption{Thrower.}
        \label{fig:ab-thrower-auc}
    \end{subfigure}
    \begin{subfigure}[h]{0.24\linewidth}
        \includegraphics[width=\linewidth]{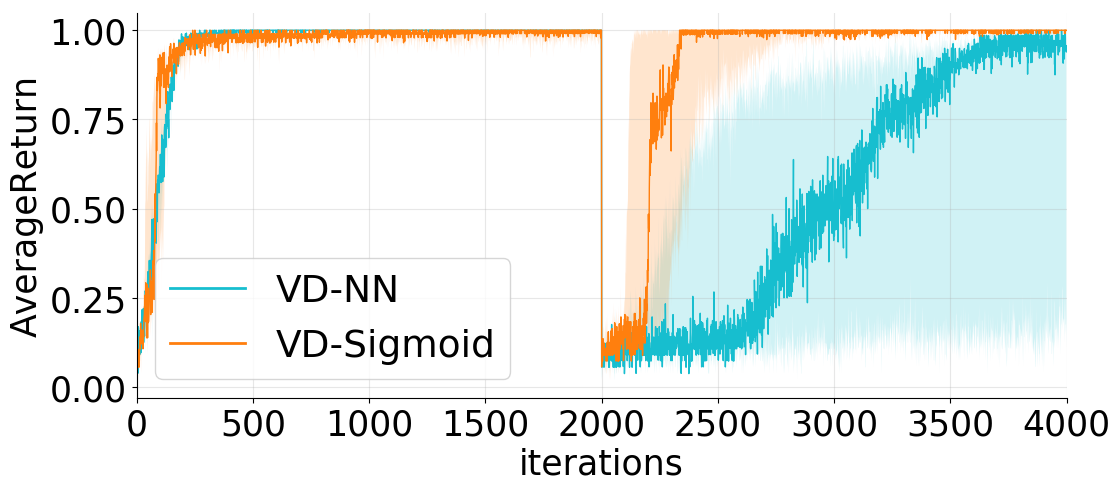}
        \caption{BiC.}
        \label{fig:ab-ball-in-cup-auc}
    \end{subfigure}
    \begin{subfigure}[h]{0.24\linewidth}
        \includegraphics[width=\linewidth]{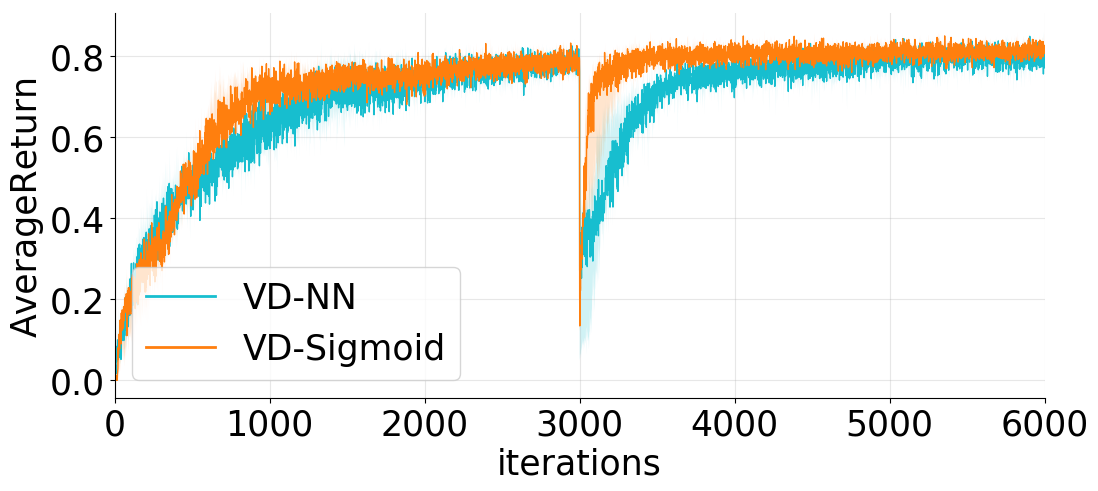}
        \caption{Slide.}
        \label{fig:ab-fetch-slide-auc}
    \end{subfigure}
    \caption{Learning curves of value dependent exploration with different mapping functions. In all figures, the environment changes are the same with the corresponding experiments shown in Figure \ref{fig:all_experiments}.}
    \label{fig:ab-all}
\end{figure*}

\textbf{Evaluation Metric}
For each algorithm, we show its median return over 5 random seeds and plot this against the number of training iterations. To evaluate the amount of actual exploration of each method, we calculate the effective entropy of each policy  after the action is clipped onto the range $[-1, 1]$.  Note that the variance by itself does not reveal the actual amount of exploration if the policy mean is much greater than 1 or much less than -1; due to the action clipping, the policy may be close to deterministic in these cases. The calculation of the entropy is given in Appendix C. A higher average entropy implies a higher degree of exploration. The clipped entropy is used only for evaluation and is not used for learning, as action-clipping is considered to be part of the environment and is unknown to the agent in the model-free setting that we are considering.

For all the environments, the learning curves and the corresponding entropy of policies learned by the different methods are plotted in Figures \ref{fig:all_experiments} and \ref{fig:all_experiments_entropy}.

% Given a random variable $x' \sim \mathcal{N}(\mu, \sigma)$, we define the clipped variable $x$ , 
% \[ x=
%     \begin{cases} 
%         x' & a \leq x' \leq b\\
%         a & x' \leq a\\
%         b & b \leq x'
%     \end{cases}
% \]

% Let $\alpha = \frac{a-\mu}{\sigma}, \beta =\frac{b-\mu}{\sigma}, Z=\Phi(\beta)-\Phi(\alpha)$, where $\Phi(x)$ and $\phi(x)$ is the cdf  and pdf of a unit Gaussian. Denote the pdf of a variable $x$ as $p(x)$.

% The entropy of $x$ is calculated as:
% $$H(x) = -P_1\log P_1 - P_2\log P_2-\int_a^b p(x)\ln p(x) dx,$$
% where $P_1 = - \int_{-\infty}^ap(x'), P_2= - \int_b^\infty p(x')$

% For a multivariate diagonal Gaussian, we apply the above definition to each action dimension and average the entropy of each dimension. 

\subsubsection{Environments details}~

\textbf{Double inverted pendulum (DIP)} In this environment, the task is to balance a double inverted pendulum. 

% At the time of $t=1000$, the center of mass of the lower pole is shifted with a horizontal offset from the geometric center.

%\ref{fig:all_experiments}.

\textbf{Control 7 DOF Arm (Thrower)} In this environment, we train a robot arm with 7 DOF to throw a ball into a basket.  This environment is denoted as Thrower. 

% The position of the basket \textbf{g} changes in each episode. The position of the robot is fixed in Stage I and moves to a new position in stage II. 
% In Stage I, the position of the robot is fixed, and in stage II, the robot moves to a new position.

\textbf{Ball in cup (BiC)} A planar actuated cup can translate in order to swing and catch a ball attached via string. This environment is denoted as BiC.  

% The changes between Stage I and Stage II are the orientation of the cup and the height of the sides of the cup.

\textbf{Fetch slide object (Slide)} In this environment, we train a Fetch robot to slide an object to the goal along a straight line. 

% The position of the goal is changed in each episode. The changes between Stage I and Stage II are the magnitude of friction coefficient between the object and the table.

All environments will change in its parameters during the half of the training. More environment details can be found in Appendix F. 

\subsubsection{Discussion}
As shown in the plots in Figure \ref{fig:all_experiments}, in stationary environments (before the environment changes), the value dependent exploration method performs about as well as the other baselines. However, when the environment changes, the average return of all methods drops at first, but the VD-Sigmoid recovers much faster. As seen from the entropy curves in Figure \ref{fig:all_experiments_entropy}, the entropy of our method makes a very large jump quickly after the environment changes and then decreases again after the policy adapts. VIME performs as well as VD-Sigmoid in the BiC experiment, as the dynamics are relatively easy to learn. However, since VIME tries to systematically explore the dynamics for all of the observed state transitions, the environmental changes lead to unstable learning and often result in poor policy performance (as shown in Figures~\ref{fig:dip-center-learning} and~\ref{fig:thrower-learning}), whereas our method leads to much more consistent policy improvements, both before and after environmental changes. The Relearn assumes knowledge of the changing point in the environment. However, re-training a model from scratch throws out all the past experiences and thus takes a longer time to learn in many cases. The fact the our model outperforms Relearn shows that past experience helps to learn new in new situations.

\subsection{Comparison of different mapping functions} \label{ablation}
 To see how much the monotonicity constraint (Theorem ~\ref{mdp_theorem_monotonic}) helps, we compare the performance of our constrained mapping function VD-Sigmoid (Equation \ref{sd-sigmoid}) to that of an unconstrained mapping function.  For the unconstrained mapping function, we use $f_\rho \colon V_\phi(s) \mapsto \sigma$ modeled by a (4,4) multi-layer perceptron (MLP) with tanh as an activation function, trained with policy gradient updates, denoted as VD-NN. We compare the learning curves of VD-NN with the one that uses our constrained function in Figure \ref{fig:ab-all}. We can see that VD-NN has slower adaptation.  The unconstrained neural network has a harder time learning an appropriate mapping function from the value to the action variance, because in high action dimensions, the relationship between the policy performance and the action variance of different action dimensions becomes complicated.  The VD-Sigmoid adapts much faster than VD-NN, showing the importance of using a function that is constrained to be monotonically decreasing. Examples of the learned mapping functions are shown in the Appendix E.

\section{Future Work}
% In this work, we propose a value-dependent exploration strategy for fast adaptation to environment changes in sparse reward environments. We analytically compute the relationship between the value and the optimal variance for a Gaussian policy. As shown in a variety of environments, our method is able to adjust the policy variance appropriately and adapt to changes much faster than traditional approaches.  

Currently our method only deals with sparse-reward environments. A possible future direction is to extend our work to dense-reward environments, allowing the agent to optimally adapt its exploration in such settings.   Since environmental changes are common in the real world, providing better exploration strategies for such cases will lead to more robust and wider application of robot learning.  

\section{Acknowledgement}
This material is based upon work supported by the United States Air Force and DARPA under Contract No. FA8750-18-C-0092.  The views and conclusions contained in this document are those of the author and should not be interpreted as representing the official policies, either expressed or implied, of any sponsoring institution, the U.S. government or any other entity.
%Further,  currently the policy's exploration is controlled by adjusting the policy variance.  In the future, we hope to explore other strategies for exploration, such as learning a more general policy for optimal exploration.

%%, as the ``success'' in such environments can be clearly defined. 



%%%%%%%%%%%%%%%%%%%%%%%%%%%%%%%%%%%%%%%%%%%%%%%%%%%%%%%%%%%%%%%%%%%%%%%%%%%%%%%%
\newpage
\bibliographystyle{IEEEtran}
\bibliography{paper}

\end{document}

% \maketitle

%; $d$ is also related to the amount of distribution shift

\section{Value to Variance Mapping}\label{monotonicity proof}
We assume that the policy is parameterized by a Gaussian distribution 
$a \sim \mathcal{N}(\mu,\,\sigma^{2})\,$. We define the sparse reward function at time $t$ as 
\[ r_t(a)=
    \begin{cases} 
        1 & l(t) \leq a \leq l(t) + w \\
        0 & \textrm{otherwise},
    \end{cases}
\]
where $w$ is the (unknown) constant that determine the width of the interval in the action space from which we get a positive reward. WLOG, assume $\mu<l(t)$, and define $d=l(t)-\mu$ as the distance from the action mean to the closest action that can get a positive reward and will change with $t$. Since $d$ is what we actually want, we will have all the variables dependent on $d$ instead of $t$. For example, we use $V_\pi(\sigma,d)$ instead of $V^t_\pi(\sigma)$. $\mu$ is the action mean which is kept fixed during our analysis. We can write the value function of a policy as a function of $\sigma$ and $d$: 
%; $d$ is also related to the amount of distribution shift
%For the remainder of this analysis, we consider optimization of the variance $\sigma$ in one iteration and fix $\mu$ and $d$. Thus, we can write the value function of a policy as a function of $\sigma$: 
\begin{equation}
V_\pi(\sigma, d) = \mathbb{E}_{a \sim \mathcal{N}(\mu,\,\sigma^{2})}[r(a, d)] \label{equ_v_expectation}
\end{equation}

With a fixed $\mu$, the optimal variance to use for the policy is the one that maximizes the value function: 
$$ \sigma^*(d) = \operatorname*{argmax}_\sigma V_\pi(\sigma, d)$$

%  A simple illustration is shown in Supplementary Figure \ref{fig:simple_illustration}.

% \begin{figure}[h]
%     \centering
%         \includegraphics[width=0.6\linewidth]{figures/setup.png}
%         \caption{Simple illustration of the sparse-reward continuous bandit setting.}
%     \label{fig:simple_illustration}
% \end{figure}

\begin{lemma} \label{sigma_best}
Under the condition that $ w, d, \sigma>0 $, the optimal variance is given by
$$\sigma^*=\sqrt{\frac{1}{2} \frac{2dw+w^2}{\ln(1+w/d)}}.$$
\end{lemma}

\begin{proof}
As a first step, we write $V_\pi$ as a function of $\sigma$ and $d$:

\begin{equation} 
\begin{split}
V_\pi(\sigma, d) & = \int_{\mu + d}^{\mu +d+w} \frac{1}{\sqrt{2\pi \sigma^2}}e^{-\frac{1}{2} (\frac{x-\mu}{\sigma})^2}dx\\
              & = \int_{d}^{d+w} \frac{1}{\sqrt{2\pi \sigma^2}}e^{-\frac{1}{2} (\frac{x}{\sigma})^2}dx \\
              & = \int_{d/\sigma}^{(d+w)/\sigma} \sqrt{\frac{1}{2\pi}}e^{-\frac{1}{2} x^2}dx 
\end{split}
\end{equation}

We can then find the optimal $\sigma$ by setting the derivative w.r.t. $\sigma$ to zero:
\begin{equation} \label{equ2}
\begin{gathered}
    \frac{\partial V}{\partial \sigma} = 0\\
    \Rightarrow \frac{d}{{\sigma^*}^2} \phi (\frac{d}{\sigma^*}) - \frac
    {d+w}{{\sigma^*}^2}\phi(\frac{d+w}{\sigma^*}) = 0 \qquad(\text{Leibniz rule})\\
    \Rightarrow \frac{\phi(\frac{d}{\sigma^*})}{\phi(\frac{d+w}{\sigma^*})} = \frac{d+w}{d}\\
\end{gathered}
\end{equation}
\begin{equation} \label{sigma}
    \Rightarrow \sigma^* = \sqrt{\frac{1}{2} \frac{2dw+w^2}{\ln(1+w/d)}},\\
\end{equation}
where $\phi(x) = \mathcal{N}(x| 0,\,1).$ \\
\end{proof}

\begin{prop} \label{prop_approx_d}
Under the condition that $ w, d, \sigma>0 $ and $d \gg w$, the optimal variance is given by $\sigma^* = d$. Further, this expression also holds as $d\to0$
\end{prop}

\begin{proof}
By applying L'Hopital's rule on equation \ref{sigma}, we get:
\begin{equation}
\begin{split}
\lim_{w/d \to 0} {\sigma^*}^2 &= \lim_{w/d \to 0}\frac{1}{2} \frac{2dw+w^2}{\ln(1+w/d)} \\
&= \lim_{w/d \to 0}\frac{1}{2} \frac{d^2(2w/d+(w/d)^2)}{\ln(1+w/d)}\\
&= \lim_{t \to 0} d^2 \Big[\frac{1}{2} \frac{2t+t^2}{\ln(1+t)}\Big] \\
&= \lim_{t \to 0} d^2 \Big[\frac{1}{2}(2+2t)(1+t)\Big] \qquad(\text{L'Hopital's rule})\\
&= d^2
\end{split}
\end{equation}
Therefore, we have that
\begin{equation}
    \lim_{w/d \to 0} \sigma^* = d
\end{equation}

The approximation also holds for very small $d$:
\begin{equation}
\begin{split}
\lim_{d \to 0} {\sigma^*}^2 &= \frac{1}{2} \frac{\lim\limits_{d \to 0} (2dw+w^2)}{\lim\limits_{d \to 0} \ln(1+w/d)} \\
&= \frac{1}{2} \frac{w^2}{\lim\limits_{d \to 0} \ln(1+w/d)}\\
&= 0
\end{split}
\end{equation}

Therefore, we have that 
$$ \lim_{d \to 0} \sigma^* = 0$$

\end{proof}

\begin{prop} 
Under the condition that $w, d, \sigma>0 $, the optimal variance $\sigma^*$ increases when the value function $V_\pi(\sigma^*, d)$ decreases.
\label{prop_decreasing}
\end{prop}

\begin{proof}
We will prove that $$ \frac{\partial \sigma^*}{\partial V} = \frac{\partial \sigma^*}{\partial d}\frac{\partial d}{\partial V} < 0 $$
Note that
\begin{equation} 
\begin{split}
\frac{\partial \sigma^*}{\partial d} & = \frac{1}{2} \frac{\frac{\partial}{\partial d}\big(\frac{1}{2} \frac{2dw+w^2}{\ln(1+w/d)}\big)}{\sqrt{\frac{1}{2} \frac{2dw+w^2}{\ln(1+w/d)}}}\\
&= \frac{1}{2} \frac{2w\ln(1+w/d) + (2dw+w^2)\frac{w}{(1+w/d)d^2}}{\sqrt{\frac{1}{2} \frac{2dw+w^2}{\ln(1+w/d)}}\ln(1+w/d)^2}\\
& > 0. 
\end{split}
\end{equation}

And 
% https://math.stackexchange.com/questions/1755149/derivative-of-error-function
\begin{equation} \label{equ_v}
\begin{split}
% \frac{\partial d}{\partial V_\pi}
V_\pi & = \int_{d/\sigma}^{(d+w)/\sigma} \sqrt{\frac{1}{2\pi}}e^{-\frac{1}{2} x^2}dx  \\
      & = \frac{1}{2} [1 + {erf}(\frac{d+w}{\sqrt{2}\sigma})] - \frac{1}{2} [1 + {erf}(\frac{d}{\sqrt{2}\sigma})] \\ 
\end{split}
\end{equation}
where erf is the error function. Recall that $erf'(x) = \frac{2}{\sqrt{\pi}}e^{-x^2}$. Differentiate on both sides w.r.t. $V_\pi$, and we get

\begin{equation} 
\begin{split}
 1 & = \frac{1}{2}\bigg[\frac{2}{\sqrt{2\pi}\sigma} e^{-(\frac{d+w}{\sqrt{2}\sigma})^2}\frac{\partial d}{\partial V_\pi} - \frac{2}{\sqrt{2\pi} \sigma} e^{-(\frac{d}{\sqrt{2}\sigma})^2}\frac{\partial d}{\partial V_\pi}\bigg] \\
& \Rightarrow \frac{\partial d}{\partial V_\pi} = \frac{\sqrt{2\pi}\sigma}{e^{-(\frac{d+w}{\sqrt{2}\sigma})^2} - e^{-(\frac{d}{\sqrt{2}\sigma})^2}}
\end{split}
\end{equation}
Since $e^{-x}$ is monotonically decreasing and $\big(\frac{d+w}{\sqrt{2}\sigma}\big)^2 > \big(\frac{d}{\sqrt{2}\sigma}\big)^2$, we have $\frac{\partial d}{\partial V_\pi} < 0$.

Thus, $$ \frac{\partial \sigma^*}{\partial V_\pi} = \frac{\partial \sigma^*}{\partial d}\frac{\partial d}{\partial V_\pi} < 0 $$\\
\end{proof}

\begin{theorem}  \label{prop_inverse_expression}
%Assuming that $w/d \to 0$ and $\sigma^*=d$, we can approximate $\sigma^*$ as:
Under the condition that $ w, d, \sigma>0 $ and $d \gg w$, the optimal variance can be written in terms of the value function as
$$\sigma^* = \frac{w}{\sqrt{2\pi e} V_\pi(\sigma^*, d)}$$
\end{theorem}

\begin{proof}
Recall from Proposition \ref{prop_approx_d} that, $\lim\limits_{w/d \to 0} \sigma^*=d$.  
We can approximate the error function in equation \ref{equ_v} with a Taylor series.  As $$\frac{d+w}{\sqrt{2}\sigma^*} = 
\frac{1+w/d}{\sqrt{2}\sigma^*/d} \approx \frac{1 + w/d}{\sqrt{2}} \approx \frac{\sqrt{2}}{2}$$ we take the Taylor series of $erf(x)$ around the point $x = \frac{\sqrt{2}}{2}$ to first order:
$$ erf(x) \approx erf(\frac{\sqrt{2}}{2}) + erf'(\frac{\sqrt{2}}{2})(x-\frac{\sqrt{2}}{2}).$$ Then from equation \ref{equ_v}, we get:
\begin{equation}
\begin{split}
V_\pi(\sigma^*, d) &= \frac{1}{2} [1 + {erf}(\frac{d+w}{\sqrt{2}\sigma^*})] - \frac{1}{2} [1 + {erf}(\frac{d}{\sqrt{2}\sigma^*})] \\ 
&= \frac{1}{2} [{erf}(\frac{d+w}{\sqrt{2}\sigma^*}) - {erf}(\frac{d}{\sqrt{2}\sigma^*})] \\ 
&\approx \frac{1}{2} \Bigg[{erf}(\frac{\sqrt{2}}{2}) + {erf}'(\frac{\sqrt{2}}{2})\Big[\frac{d+w}{\sqrt{2}\sigma^*} - \frac{\sqrt{2}}{2}\Big]-{erf}(\frac{\sqrt{2}}{2}) - {erf}'(\frac{\sqrt{2}}{2})\Big[\frac{d}{\sqrt{2}\sigma^*}  - \frac{\sqrt{2}}{2}\Big]\Bigg]\\
&= \frac{1}{2}erf'(\frac{\sqrt{2}}{2}) \frac{w}{\sqrt{2}\sigma^*}\\
&= e^{-\frac{1}{2}} \frac{w}{\sqrt{2\pi}\sigma^*}\\
\end{split}
\end{equation}
For the last step in the above equation, recall that $erf'(x) = \frac{2}{\sqrt{\pi}}e^{-x^2}$. Rearranging the terms in the above equation, we have:
$$\sigma^* = \frac{w}{\sqrt{2\pi e} V_\pi(\sigma^*, d)}$$
\end{proof}

\begin{prop}
When using policy gradient methods with a batch size of $n$, assuming $V_\pi(\sigma, d)$ is close to zero, if we increase the value function $V_\pi$ by $\delta$, the probability of getting at least one positive reward within the batch increases by $n \delta$.
\end{prop}

\begin{proof}
Since we define the reward to be either $0$ or $1$, the probability of sampling a positive reward from our policy is:
%, we separate the parts where we get reward $1$ and the parts where we get reward $0$:
\begin{equation}
    \begin{split}
        P(r_\pi(a) > 0)  & = \int_{\mu + d}^{\mu +d+w} \frac{1}{\sqrt{2\pi \sigma^2}}e^{-\frac{1}{2} (\frac{x-\mu}{\sigma})^2}dx \\
        % & + 0\int_{-\infty}^{\mu +d} \frac{1}{\sqrt{2\pi \sigma^2}}e^{-\frac{1}{2} (\frac{x-\mu}{\sigma})^2}dx + 0\int_{\mu + d + w}^{+\infty} \frac{1}{\sqrt{2\pi \sigma^2}}e^{-\frac{1}{2} (\frac{x-\mu}{\sigma})^2}dx\\
        &= V_\pi(\sigma, d)
    \end{split}
\end{equation}

Thus the probability of getting at least one positive reward in a batch size of $n$ is $ P^+_n = 1-(1-V_\pi(\sigma, d))^n$. As the Taylor series of $(1-x)^n$ at point $x=0$ is $\sum\limits_{i=0}^\infty \binom{n}{i}(-x)^i$, we approximate $P^+_n$ at $V_\pi(\sigma) \to 0$ to the second term of the Taylor series:
$$P^+_n = 1-(1-V_\pi(\sigma, d))^n \approx 1 - (1 - n V_\pi(\sigma, d)) = n V_\pi(\sigma, d).$$
Thus, increasing $V_\pi(\sigma, d)$ by $\delta$ will increase $P^+_n$ by $n \delta$.

\end{proof}

\section{Monotonicity in MDP}\label{mdp proof}
Since we are interested in the relationship between the action variance and the value function at the current state $s_t$ which is fixed, we omit $s_t$ in the following analysis. For example, the $Q$ function will be written as a function of only action $a$. As a first step, we assume that the initial Q function (before any environment changes) is a bounded function in the action dimension, i.e.
$$ Q^\pi_0(a) = 0 \text{ if } a < l_0   \text{ or } l_0 + w < a.$$
Additionally,  as all rewards can be normalized to $[0, 1]$, we can assume $\forall a, Q^\pi_0(a)>0$ WLOG. As the environment changes, Q function will also change. Here we assume that when the change in the environment is small, Q function changes only through translation without shape changing. Thus, we define
$$ Q_t^\pi(a) = 0 \text{ if }  a< l(t) \text{ or } l(t) + w < a.$$
Define $d = l(t) - \mu$ as the distance from $\mu$ to the closest action where Q is not zero at time step t. Although $d$ is implicitly dependent on $t$, we simplify the notation by dropping $t$ in $d$, since only the distance matters to our analysis. As $Q_t$ is only a translation of $Q_0, Q_t(a) = Q^\pi_d(a) = Q^\pi_0(a-(l(t)-l_0))$. 
By definition of the value function, we have $$V_\pi(d) = \int \pi(a) Q^\pi_d(a)da,$$ where $V_\pi(d)$ is the value function given a certain $d$. Since the policy is parameterized as a normal distribution $\mathcal{N}(\mu, \sigma)$ and for this analysis, let us assume $\mu$ is fixed so that we can independently analyze the effect of the policy variance and return, we can write the value function as a function of $\sigma$ and $d$: $V(\sigma, d) = V_\pi(d)= \int p(a| \mu, \sigma) Q^\pi_d(s, a),$ where $p(\cdot | \mu, \sigma)$ is the density of the Gaussian distribution.

\begin{lemma} \label{mdp_prop_sigma_bound}
Under the condition that $d, w, \sigma >0$, if the mean action of the policy $\mu$ is fixed, the optimal variance $\sigma^* \in (d, d+w)$. 
\end{lemma}
\begin{proof}
Let us calculate the derivative of $V$ w.r.t. $\sigma$:

\begin{equation}
\begin{split}
\frac{\partial V(\sigma, d)}{\partial \sigma} &= \frac{\partial}{\partial \sigma}\int_{\mu+d}^{\mu+d+w} p(a|\sigma) Q^\pi_d(a)da \\
&=\int_{\mu+d}^{\mu+d+w} \frac{\partial}{\partial \sigma}p(a|\sigma) Q^\pi_d(a)da \\
&=\int_{\mu+d}^{\mu+d+w} \frac{1}{\sqrt{2\pi} \sigma^2}e^{-\frac{(a-\mu)^2}{2\sigma^2}}\big[\frac{(a-\mu)^2}{\sigma^2} -1\big] Q^\pi_d(a)da \\
\label{eqn:mdp_dsigma}
\end{split}
\end{equation}
Since $\sigma^*$ is optimal, we know that $\frac{\partial V(\sigma)}{\partial \sigma}\Big|_{\sigma=\sigma^*} =0$. However, when $\sigma < d$, $\frac{(a-\mu)^2}{\sigma^2} -1 > 0, \forall a \in [\mu+d, \mu+d+w]$. Thus, when $\sigma<d$, all terms being integrated in Equation \ref{eqn:mdp_dsigma} are larger than zero and $\frac{\partial V(\sigma)}{\partial \sigma}>0$ . Similarly, when $\sigma > d+w$, $\frac{\partial V(\sigma)}{\partial \sigma}<0$. 

Since $\frac{\partial V(\sigma)}{\partial \sigma}|_{\sigma < d} > 0$ and $\frac{\partial V(\sigma)}{\partial \sigma}|_{\sigma > d+w} < 0$, from the intermediate value theorem we know that there exists $\sigma^* \in (d, d+w)$, such that$\frac{\partial V(\sigma)}{\partial \sigma}|_{\sigma=\sigma^*} =0.$
\end{proof}

\begin{theorem} \label{mdp_theorem_monotonic}
Under the condition that $d, w, \sigma >0$, if $w$ and $d$ satisfies that $w/d < \sqrt{3} -1$ then the optimal variance $\sigma^*$ increases when $V(\sigma^*, d)$ decreases.
\end{theorem}
\begin{proof}
We prove that $$ \frac{\partial \sigma^*}{\partial V(\sigma^*, d)} = \frac{\partial \sigma^*}{\partial d}\frac{\partial d}{\partial V(\sigma^*, d)} <0.$$

First, we prove that $\frac{\partial \sigma^*}{\partial d} > 0$ . For any $\sigma$, 
\begin{equation}
\begin{split}
V(\sigma, d) & = \int_{\mu + d}^{\mu+d+w} p(a; \mu, \sigma) Q^\pi_d(a) da \\
&= \int_{\mu + d}^{\mu+d+w} p(a; \mu, \sigma) Q^\pi_0(a-l(d)+l_0) da\\
&= \int_{\mu + d}^{\mu+d+w} p(a; \mu, \sigma) Q^\pi_0(a-\mu-d+l_0) da\\
&= \int_0^w p(t+d+\mu; \mu, \sigma) Q^\pi_0(t+l_0) dt\qquad(\text{Let $t = a-\mu-d$}) \\
&= \int_0^w \frac{1}{\sqrt{2\pi}\sigma}e^{-\frac{(t+d)^2}{2\sigma^2}}Q^\pi_0(t+l_0) dt \\
\end{split}
\end{equation}
The derivative of the value function w.r.t the optimal variance should equal to zero:
\begin{equation} \label{mdp_partial_sigma}
\begin{split}
\frac{\partial V(\sigma, d)}{\partial \sigma }\Big|_{\sigma=\sigma^*} & =\int_0^w \frac{1}{\sqrt{2\pi}(\sigma^*)^2}\Big[\frac{(t+d)^2}{(\sigma^*)^2} - 1\Big]e^{-\frac{(t+d)^2}{2(\sigma^*)^2}}Q^\pi_0(t+l_0) dt  = 0
\end{split}
\end{equation}
Taking the derivative w.r.t. $d$ on equation (\ref{mdp_partial_sigma}) and we get:
\begin{equation} \label{mdp_three_term}
\begin{split}
&\int_0^w -\frac{1}{\sqrt{2\pi}}\frac{2}{(\sigma^*)^3} \Big[\frac{(t+d)^2}{(\sigma^*)^2} - 1\Big]e^{-\frac{(t+d)^2}{2(\sigma^*)^2}}Q^\pi_0(t+l_0) dt \qquad(\text{I})\\
+& \int_0^w \frac{1}{\sqrt{2\pi}}\frac{1}{(\sigma^*)^2}  \big[-(\frac{t+d}{\sigma^*}\frac{\sigma^*-(t+d)\frac{\partial \sigma^*}{\partial d}}{(\sigma^*)^2})\big]\Big[\frac{(t+d)^2}{(\sigma^*)^2} - 1\Big]e^{-\frac{(t+d)^2}{2(\sigma^*)^2}}Q^\pi_0(t+l_0) dt \qquad(\text{II})\\
+& \int_0^w \frac{1}{\sqrt{2\pi}}\frac{1}{(\sigma^*)^2}  \Big[2\frac{t+d}{\sigma^*}\frac{\sigma^*-(t+d)\frac{\partial \sigma^*}{\partial d}}{(\sigma^*)^2}\Big]e^{-\frac{(t+d)^2}{2(\sigma^*)^2}}Q^\pi_0(t+l_0) dt  \qquad(\text{III})\\
=&0
\end{split}
\end{equation}
Compare term (I) in equation (\ref{mdp_three_term}) to equation (\ref{mdp_partial_sigma}), we know that term (I) equals to zero. After organization, we re-write equation (\ref{mdp_three_term}) as:
\begin{equation}
\begin{split}
&\int_0^w \frac{1}{\sqrt{2\pi}}\frac{1}{(\sigma^*)^2} \Big(\frac{t+d}{\sigma^*}\Big)\Big[3-\Big(\frac{t+d}{\sigma^*}\Big)^2\Big]e^{-\frac{(t+d)^2}{2(\sigma^*)^2}}Q^\pi_0(t+l_0)\frac{1}{\sigma^*} dt \\
=&\int_0^w \frac{1}{\sqrt{2\pi}}\frac{1}{(\sigma^*)^2} \Big(\frac{t+d}{\sigma^*}\Big)\Big[3-\Big(\frac{t+d}{\sigma^*}\Big)^2\Big]e^{-\frac{(t+d)^2}{2(\sigma^*)^2}}Q^\pi_0(t+l_0)\Big(\frac{t+d}{(\sigma^*)^2}\Big)dt \frac{\partial \sigma^*}{\partial d}\\
\end{split}
\end{equation}
From Lemma \ref{mdp_prop_sigma_bound}, we know that $\sigma^* \in (d, d+w)$. Combine this with the assumption that $w/d < \sqrt{3} -1$, we have 
$$ \frac{1}{1+w/d}=\frac{d}{w+d} \leq \frac{t+d}{\sigma^*} \leq \frac{w+d}{d} = 1+ \frac{w}{d}<\sqrt{3}, \forall t \in [0, w]$$
Now we know that
$$\Big[3-\Big(\frac{t+d}{\sigma^*}\Big)^2\Big]>0$$ 
 and that all terms being integrated are positive. Therefore, $\frac{\partial \sigma^*}{\partial d}>0$.

Second, we prove that $\frac{\partial d}{\partial V(\sigma^*, d)} <0$. We directly calculate the derivative:
\begin{equation} \label{mdp_partial_v_partial_d}
\begin{split}
\frac{\partial V(\sigma^*, d)}{\partial d} &= \frac{\partial}{\partial d}\int_0^w \frac{1}{\sqrt{2\pi}\sigma^*}e^{-\frac{(t+d)^2}{2(\sigma^*)^2}}Q^\pi_0(t+l_0) dt \\
&=\int_0^w -\frac{1}{\sqrt{2\pi}(\sigma^*)^2}e^{-\frac{(t+d)^2}{2(\sigma^*)^2}}Q^\pi_0(t+l_0) dt \frac{\partial \sigma^*}{\partial d}\\
& - \int_0^w \frac{1}{\sqrt{2\pi}\sigma^*}e^{-\frac{(t+d)^2}{2(\sigma^*)^2}}\Big[\frac{t+d}{(\sigma^*)^2} - \frac{(t+d)^2}{(\sigma^*)^3}\frac{\partial \sigma^*}{\partial d}\Big]Q^\pi_0(t+l_0) dt \\
&= \int_0^w \frac{1}{\sqrt{2\pi}(\sigma^*)^2}e^{-\frac{(t+d)^2}{2(\sigma^*)^2}}\Big[\frac{(t+d)^2}{(\sigma^*)^2}-1\Big]Q^\pi_0(t+l_0) dt \frac{\partial \sigma^*}{\partial d} \qquad(\text{I})\\
&+ \int_0^w -\frac{1}{\sqrt{2\pi}\sigma^*}e^{-\frac{(t+d)^2}{2(\sigma^*)^2}}\frac{t+d}{(\sigma^*)^2}Q^\pi_0(t+l_0)dt \qquad(\text{II}) \\
\end{split}
\end{equation}
From equation (\ref{mdp_partial_sigma}), we know that the term (I) in equation (\ref{mdp_partial_v_partial_d}) equals to zero. And $\forall t \in [0, w]$, term (II) in equation (\ref{mdp_partial_v_partial_d}) is less or equal to zero. Thus, $\frac{\partial V(\sigma^*, d)}{\partial d}<0$ and $\frac{\partial d}{\partial V(\sigma^*, d)} = \frac{1}{\frac{\partial V(\sigma^*, d)}{\partial d}} < 0$.
\end{proof}

\begin{definition} (Highest Density Region)
Let $f(x)$ be the density function of a random variable $X$. Then the $(1-\alpha)$ HDR is the subset $R(f_\alpha)$ of the sample space of $X$ such that 
$$ R(f_\alpha)  =\{x:f(x) \leq f_\alpha\}$$
where $f_\alpha$ is the largest constant such that $$ P(X\in R(f_\alpha)) \leq 1-\alpha$$
\end{definition}
\begin{prop}
Assume that $d, w, \sigma >0$ and $(1-\alpha)$ HDR of $Q^\pi_d$ can be bounded by $[l(d), r(d)]$, $0<\alpha<1$. Let $\hat{Q^\pi_d}$ be the clipped version of $Q^\pi_d$, where
\[ \hat{Q}^\pi_d(a)=
    \begin{cases} 
        Q^\pi_d(a) &  l(d) \leq a \leq r(d)\\
        0 & otherwise\\
    \end{cases}. 
\]
Let $\sigma^*$ be the best variance of $Q(d)$ and $\hat{\sigma}^*$ be the corresponding best variance for $\hat{Q}^\pi(\sigma)$. Then $$ |V(\sigma^*,d) - V(\hat{\sigma}^*, d)| \leq 1-\alpha$$
\end{prop}
\begin{proof}
Define $\hat{V}(\sigma ,d) = \int p(a; \mu, \sigma^*) \hat{Q}^\pi_d(a) da$. Since $\hat{Q}^\pi_d(a) \leq Q^\pi_d(a)$, we have that $\hat{V}(\sigma, d) \leq V(\sigma, d)$. Since $\sigma^*$ maximizes $V(\sigma, d)$, $$ V(\hat{\sigma}^*, d)- V(\sigma^*,d) \leq1-\alpha.$$ So now we only need to prove that $$V(\sigma^*,d)-V(\hat{\sigma}^*, d) \leq 1-\alpha$$ 
\begin{equation*}
\begin{split}
V(\sigma^*,d) &=  \int_{a\in [l(d), r(d)]} p(a; \mu, \sigma^*) Q^\pi_d(a) da + \int_{a\notin [l(d), r(d)]} p(a; \mu, \sigma^*) Q^\pi_d(a) da\\
&=\int_{a\in [l(d), r(d)]} p(a; \mu, \sigma^*) Q^\pi_d(a) da\\
&~~+ \int_{a\notin [l(d), r(d)]} p(a; \mu, \hat{\sigma}^*)da \int_{a\notin [l(d), r(d)]}Q^\pi_d(a) da \qquad\text{($Q_d$ is non-negative)}\\ 
&\leq \int_{a\in [l(d), r(d)]} p(a; \mu, \sigma^*) Q^\pi_d(a) da +  1 \cdot (1-\alpha)\\ 
&\leq \int_{a\in [l(d), r(d)]} p(a; \mu, \sigma^*) \hat{Q}^\pi_d(a) da +   (1-\alpha)\\ 
&=\hat{V}(\sigma^*, d) +   (1-\alpha)\\ 
&\leq \hat{V}(\hat{\sigma}^*, d) +   (1-\alpha)  \qquad\text{($\hat{\sigma}^*$ maximizes $\hat{V}(\sigma, d)$)}\\
&\leq V(\hat{\sigma}^*, d) +   (1-\alpha)\\ 
\end{split}
\end{equation*}

\end{proof}

\newpage
\section{Entropy of Clipped Gaussian}
Given a random variable $x' \sim \mathcal{N}(\mu, \sigma)$, we define the clipped variable $x$:
\[ x=
    \begin{cases} 
        x' & a \leq x' \leq b\\
        a & x' \leq a\\
        b & b \leq x'
    \end{cases}
\]

Let $\alpha = \frac{a-\mu}{\sigma}, \beta =\frac{b-\mu}{\sigma}, Z=\Phi(\beta)-\Phi(\alpha)$, where $\Phi(x)$ and $\phi(x)$ are the cdf  and pdf of a unit Gaussian respectively. Denote the pdf of a variable $x$ as $p(x)$.

The entropy of $x$ is calculated as:
$$H(x) = -P_1\ln P_1 - P_2\ln P_2-\int_a^b p(x)\ln p(x) dx,$$
where $P_1 = \int_{-\infty}^ap(x')dx', P_2=\int_b^\infty p(x')dx'$
\begin{equation}
\begin{split}
-\int_a^b p(x) \ln p(x)dx 
&= -\int_a^b \frac{1}{\sqrt{2\pi\sigma^2}} e^{-\frac{(x-\mu)^2}{2\sigma^2}} \big[-\ln(\sqrt{2\pi\sigma^2}) - \frac{(x-\mu)^2}{2\sigma^2}\big]dx\\
&=-\int_\frac{a-\mu}{\sigma}^\frac{b-\mu}{\sigma} \frac{1}{\sqrt{2\pi}}e^{-\frac{x^2}{2}} \big[-\ln(\sqrt{2\pi\sigma^2}) - \frac{x^2}{2}\big]dx\\
&=\ln(\sqrt{2\pi\sigma^2}) \int_\alpha^\beta\phi(x)dx - \frac{1}{2\sqrt{2\pi}}\int_\alpha^\beta x de^{-\frac{x^2}{2}}\\
&= Z\ln(\sqrt{2\pi\sigma^2}) - 
\frac{1}{2\sqrt{2\pi}} \Big[ xe^{-\frac{x^2}{2}}\Big|_\alpha^\beta - \int_\alpha^\beta e^{-\frac{x^2}{2}}dx\Big]\\
&= Z\ln(\sqrt{2\pi\sigma^2}) - 
\frac{\beta\phi(\beta) - \alpha\phi(\alpha)- Z}{2} \\
&= Z\ln(\sqrt{2\pi e}\sigma) + 
\frac{\alpha\phi(\alpha) - \beta\phi(\beta)}{2}\\
\end{split}
\end{equation}
Thus
$$ H(x) = Z\ln(\sqrt{2\pi e}\sigma) + 
\frac{\alpha\phi(\alpha) - \beta\phi(\beta)}{2}  
-P_1\ln P_1 - P_2\ln P_2$$

For a multivariate diagonal Gaussian, we apply the above definition to each dimension and average the entropy of each dimension.  In our environments, actions are normalized to be in the range [-1,1], so we use $a=-1$, $b=1$. 
\newpage
\section{Convergence Analysis on Continuous Bandit}
% Consider the continuous bandit problem as illustrated in Supplementary Figure \ref{fig:simple_illustration}. 
Consider the continuous bandit problem in Appendix A. First, we fix the policy mean $\mu$ and learn the parameter $w$ through gradient descent. At the same time, the policy variance is being updated through Theorem \ref{prop_inverse_expression}. This is the same value dependent exploration as described in Section IV.A, except that we fix the policy mean to better understand the convergence of the algorithm. Specifically, we use ADAM optimizer with a batch size of $10000$ to learn $w$, where the expected reward is estimated by sampling.  We empirically show that, in this case, the learned $w$ converges to its true value and the calculated policy variance converges to the optimal variance calculated by Lemma \ref{sigma_best}. 

The results are shown in Supplementary Figure \ref{fig:converge}. After about 100 iterations, the learned $w$ converges to its true value, and the action variance, calculated by Theorem \ref{prop_inverse_expression}, converges to the optimal variance suggested by Lemma \ref{sigma_best}. These convergence results show that, first, it is possible to learn $w$ and thus the optimal mapping function by policy gradient descent. Second, despite all the approximations that we have made during the derivation of Theorem \ref{prop_inverse_expression}, the calculated variance still converges to the optimal variance.
\begin{figure}[h!]
    \centering
    \begin{subfigure}[h]{0.45\linewidth}
        \includegraphics[width=\linewidth]{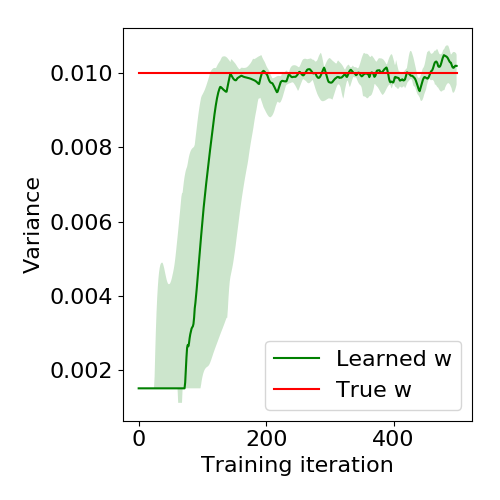}
        \caption{}
        \label{fig:converge_w}
    \end{subfigure} \hspace{0.05\textwidth}
    \begin{subfigure}[h]{0.45\linewidth}
        \includegraphics[width=\linewidth]{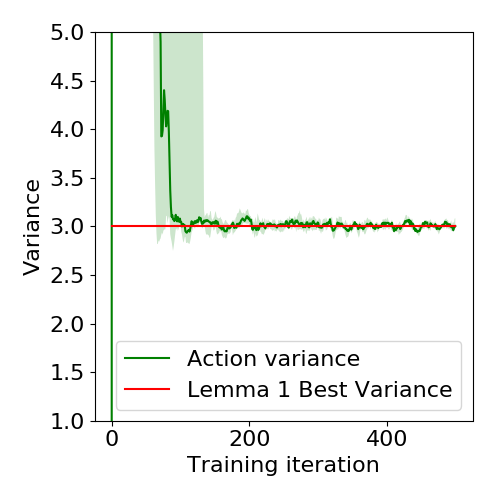}
        \caption{}
        \label{fig:converge_sigma}
    \end{subfigure}
    \caption{ In Figure \ref{fig:converge_w},  we show the convergence of the learned $w$ converges to its true value during training time. \ref{fig:converge_sigma} shows the convergence of the action variance (green line), which is calculated using Theorem \ref{prop_inverse_expression} during training time, to the optimal variance (red line) calculated by Lemma \ref{sigma_best}.}
    \label{fig:converge}
\end{figure}

Finally, we show that the empirical results of convergence shown above are also true for different values of $d$. These are shown in Supplementary Figure \ref{fig:converge_all}. Here we run the experiments of learning $w$ and policy variance as mentioned above for 500 iterations and plot the final converged $w$ against its true values in Supplementary Figure \ref{fig:converge_all_w}. We also plot the final converged policy variance against the optimal variance calculated by Lemma \ref{sigma_best} in Supplementary Figure \ref{fig:converge_all_sigma}. Note that, as $d$ gets larger, more samples are needed for the variance to converge to the optimal variance. That explains why the action variance slightly deviates from the optimal variance in Supplementary Figure \ref{fig:converge_all_sigma} as $d$ increases. However, in all cases, the learned $w$ still converges to its true value, and the action variance converges to the optimal variance.

\begin{figure}[h]
    \centering
    \begin{subfigure}[h]{0.45\linewidth}
        \includegraphics[width=\linewidth]{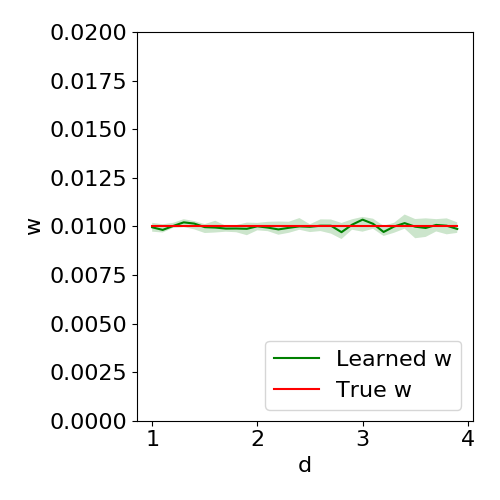}
        \caption{}
        \label{fig:converge_all_w}
    \end{subfigure} \hspace{0.05\textwidth}
    \begin{subfigure}[h]{0.45\linewidth}
        \includegraphics[width=\linewidth]{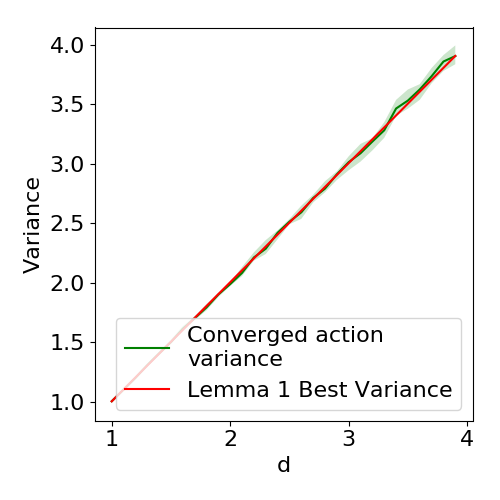}
        \caption{}
        \label{fig:converge_all_sigma}
    \end{subfigure}
    \caption{Similar to Figure \ref{fig:converge}, In Figure \ref{fig:converge_all_w},  we show the final values of the learned $w$(green line) and its true value with different $d$. \ref{fig:converge_all_sigma} shows the final action variance (green line), which is calculated using Theorem \ref{prop_inverse_expression}, to the optimal variance (red line) calculated by Lemma \ref{sigma_best}.}
    \label{fig:converge_all}
\end{figure}

For experiments shown in Supplementary Figures \ref{fig:converge} and \ref{fig:converge_all}, we set $\mu=0$ and $w=0.01$. For experiments shown in Supplementary Figure~\ref{fig:converge}, we used $d=3$, and for Supplementary Figure~\ref{fig:converge_all}, we vary $d$. All the experiments are run for 20 random seeds. 
\begin{figure}[h]
    \centering
    \begin{subfigure}[h]{0.45\linewidth}
        \includegraphics[width=\linewidth]{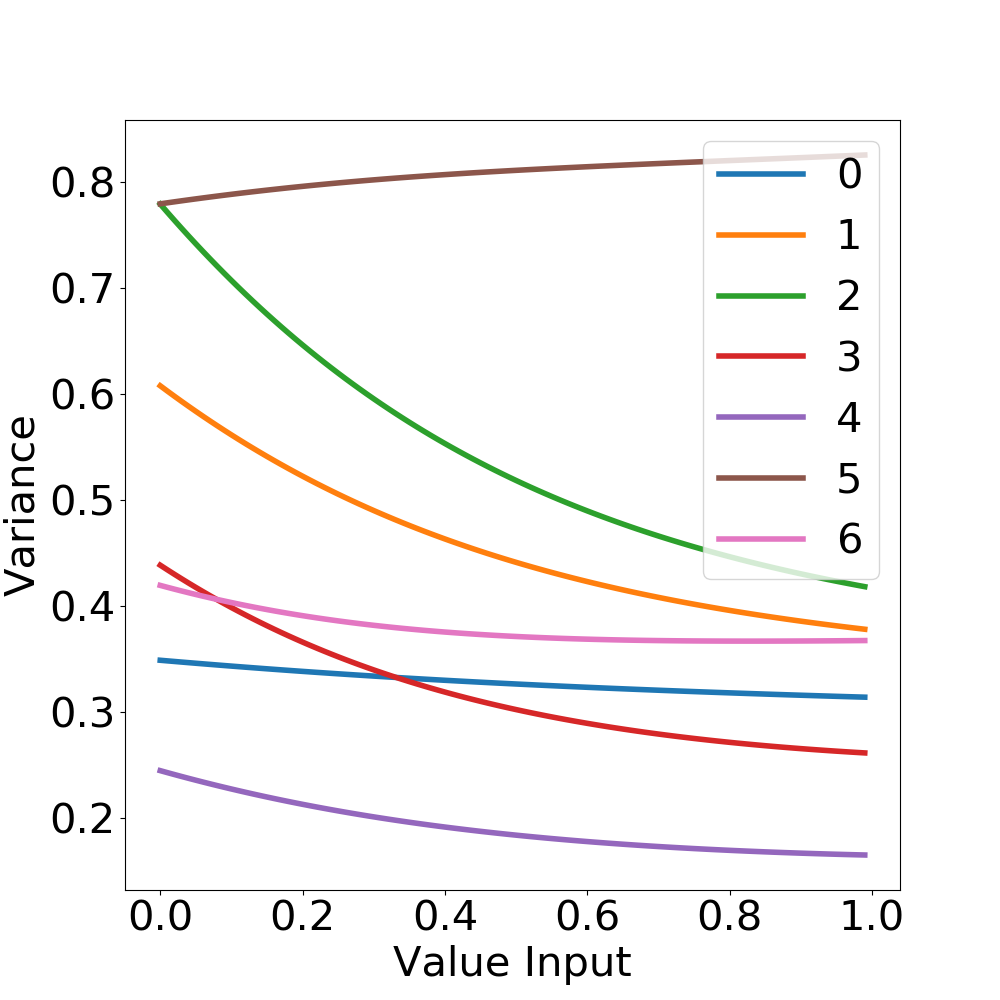}
        \label{fig:mapping-sigmoid}
        \caption{Mapping function learned by SD-Sigmoid.}
    \end{subfigure} \hspace{0.05\textwidth}
    \begin{subfigure}[h]{0.45\linewidth}
        \includegraphics[width=\linewidth]{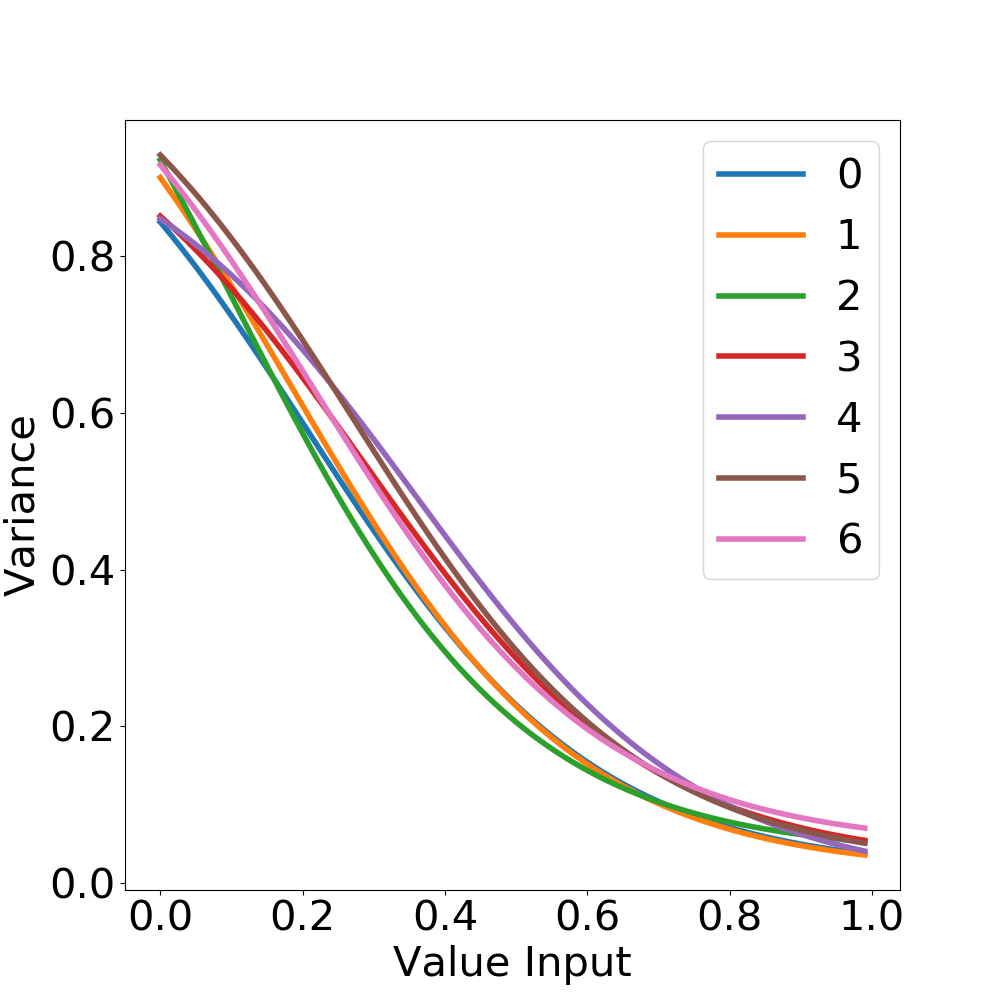}
        \label{fig:mapping-nn}
        \caption{Mapping function learned by SD-NN.}
    \end{subfigure}
    \caption{Mapping functions $f_\rho$ learned by value dependent exploration methods in the Thrower environment at time $t=2000$, right before the environment changes. Each curve stands for the mapping for one action dimension.}
    \label{fig:mapping-all}
\end{figure}
\section{Mapping Function Visualization}
We plot in Supplementary Figure \ref{fig:mapping-all} an example of the learned mapping functions, after the algorithms converge in the first stage but before the environment changes. We can see that, SD-NN does not automatically learn a strictly decreasing function and thus uses sub-optimal variances according to Theorem \ref{mdp_theorem_monotonic}.

\section{Experiment Details}
In the continuous bandit example in Section IV.A, we use $\alpha = 0.05$ and $\lambda=0.01$. For ADAM, we use $\beta_1=0.9$ and $\beta_2=0.99$, $\epsilon=1e^{-8}$, and the learning rate is $0.01$. The batch size is $128$.

In the manipulation environments in Section IV.B, all experiments use a batch size of $4000$ and a learning rate of $0.01$. For the BiC environment, we use an episode length of $500$. For other environments, we use a maximum episode length of $100$. 

For the VIME baseline, the dynamic model is estimated with a Bayesian neural network (BNN) as proposed by Houthooft, et. al.~\cite{Houthooft2016vime}. The BNN we use has one hidden layer of 32 units with ReLU non-linearity. The number of samples used for approximating the variational lower bound is 10. The size of the replay buffer is 100,000. We use second order update for the BNN with the step size $\lambda$ set to 0.01. These hyper-parameters are the ones given in \cite{Houthooft2016vime} and taken from the publicly available VIME implementation at \cite{vime_code}.
\subsection{Environment Details}
\textbf{Double inverted pendulum (DIP)} In this environment, the task is to balance a double inverted pendulum. The reward function we use is 
$$r(\textbf{x}) = \sum_{t=t_0}^T \mathbbm{1}\{\theta_{1, t}\leq 0.15 \text{ and }  \theta_{2,t} \leq 0.15 \text\} / (T-t_0)$$
where $\theta_{1,t}, \theta_{2,t}$ are angles of the two joints relative to the upright configuration. $t_0$ is the time where we start to count the number of times that the pendulums are kept upright, which is set to 10. At the time of $t=1000$, the center of mass of the lower pole is shifted with a horizontal offset from the geometric center. The shift amount is 0.02m in our experiment.

%\ref{fig:all_experiments}.

\textbf{Control 7 DOF Arm (Thrower)} In this environment, we train a robot arm with 7 DOF to throw a ball into a basket.  This environment is denoted as Thrower. The position of the basket \textbf{g} changes in each episode. The reward function we use is 
$$r(\textbf{x}) =\max(1-\norm{\textbf{x} - \textbf{g}}_2, 0) $$ where \textbf{x} is the position where the ball hits the ground. In Stage I, the position of the robot is fixed, and in stage II, the robot moves to a new position.  At $t=2000$, the position of the robot arm shifts horizontally from 0 m to -0.6 m.

\textbf{Ball in cup (BiC)} A planar actuated cup can translate in order to swing and catch a ball attached via string. This environment is denoted as BiC.  The reward function we use is
$$r(\textbf{x})=\mathbbm{1}\{\textrm{dist}(ball,target) < \textrm{size}(target) - \textrm{size}(ball)\}$$
The changes between Stage I and Stage II are the orientation of the cup and the height of the sides of the cup. At $t=2000$, the orientation of the cup changes from \ang{-45} degrees to \ang{45} degrees and the height of the sides of the cup changes from 0.12 m to 0.20 m.

\textbf{Fetch slide object (Slide)} In this environment, we train a Fetch robot to slide an object to the goal along a straight line. The position of the goal is changed in each episode. The reward function we use is
$$r(\textbf{x}) =\mathbbm{1}\{\textrm{dist}(ball,target) < 0.5\} * 1/(1+e^{50\times\textrm{dist}-5})$$
The changes between Stage I and Stage II are the magnitude of friction coefficient.  At $t=2000$, the friction coefficient of the table changes from 0.1 degrees to 0.18.
% 1d Gaussian example
% sd-sigmoid initialized values

\small
\bibliographystyle{unsrt}
\bibliography{supplement}